%% file: main.tex
\documentclass[10pt,twocolumn,letterpaper]{article}

\usepackage[pagenumbers]{iccv} % To force page numbers, e.g. for an arXiv version


\usepackage[table]{colortbl}
\usepackage{multirow}
\definecolor{iccvblue}{rgb}{0.21,0.49,0.74}
\usepackage[pagebackref,breaklinks,colorlinks,allcolors=iccvblue]{hyperref}

\def\paperID{5532} % *** Enter the Paper ID here
\def\confName{ICCV}
\def\confYear{2025}

\title{Faster and Better 3D Splatting via Group Training}

\author{
Chengbo Wang\textsuperscript{1,3,4,5} \qquad
Guozheng Ma\textsuperscript{2} \qquad
Yifei Xue\textsuperscript{1,3,4,5} \qquad
Yizhen Lao\textsuperscript{1,4,5,$\dagger$} \\
\textsuperscript{1}Hunan University 
\quad 
\textsuperscript{2}Nanyang Technological University
\\
\textsuperscript{3}DaHe.AI 
\qquad 
\textsuperscript{4}Lushan Innovation Lab 
\\
\textsuperscript{5}
Key Laboratory of Digital Culture Smart Design Technology, Minstry of Culture and Tourism
\\
{\tt\small \{wangchb, yizhenlao\}@hnu.edu.cn, GUOZHENG001@e.ntu.edu.sg, iflyhsueh@gmail.com}
}
\begin{document}
\maketitle
\renewcommand{\thefootnote}{\relax}
\footnotetext{
\textsuperscript{$\dagger$} Corresponding author: Yizhen Lao (\tt yizhenlao@hnu.edu.cn)
}
\input{sec/0_Abstract}
\input{figures/0_teaser}

\input{sec/1_Intro}
\input{sec/2_Related}
\input{sec/3_Method}

\input{sec/4_Experiments}

\input{sec/5_Conclution}

\clearpage
\input{sec/X_acknowledgment}

{
    \small
    \bibliographystyle{ieeenat_fullname}
    \bibliography{main}
}

\input{sec/X_suppl}

\end{document}

%% file: sec/0_Abstract.tex
\begin{abstract}
3D Gaussian Splatting (3DGS) has emerged as a powerful technique for novel view synthesis, demonstrating remarkable capability in high-fidelity scene reconstruction through its Gaussian primitive representations. However, the computational overhead induced by the massive number of primitives poses a significant bottleneck to training efficiency. To overcome this challenge, we propose Group Training, a simple yet effective strategy that organizes Gaussian primitives into manageable groups, optimizing training efficiency and improving rendering quality. This approach shows universal compatibility with existing 3DGS frameworks, including vanilla 3DGS and Mip-Splatting, consistently achieving accelerated training while maintaining superior synthesis quality. Extensive experiments reveal that our straightforward Group Training strategy achieves up to 30\% faster convergence and improved rendering quality across diverse scenarios. Project Website: \url{https://chengbo-wang.github.io/3DGS-with-Group-Training/}. 

\end{abstract}

%% file: figures/0_teaser.tex
 \begin{figure}[!b]
    \vspace{-20pt}
    \centering
    \scalebox{0.93}{ % 0.9 scale
    \begin{subfigure}{1.\linewidth}
        \centering
        \includegraphics[trim={0.cm 0 0.cm 0}, clip, width=\linewidth]{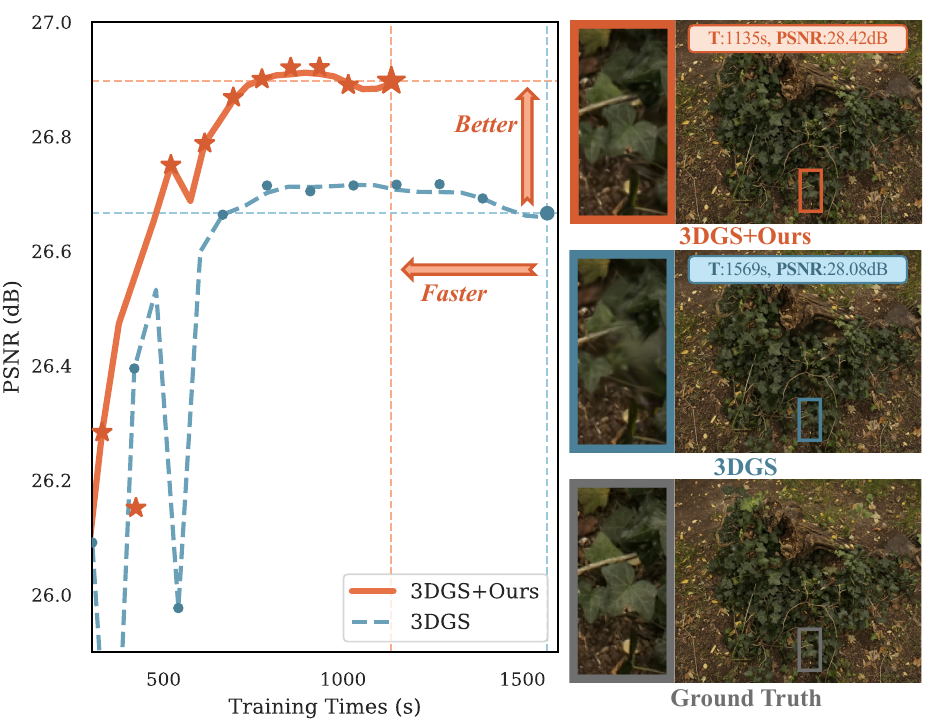}
        \caption{Reconstruction of the “Stump”\cite{barron2022mipnerf360} scene by 3DGS\cite{kerbl3Dgaussians}.}
        \includegraphics[trim={0.cm 0 0.cm 0}, clip, width=\linewidth]{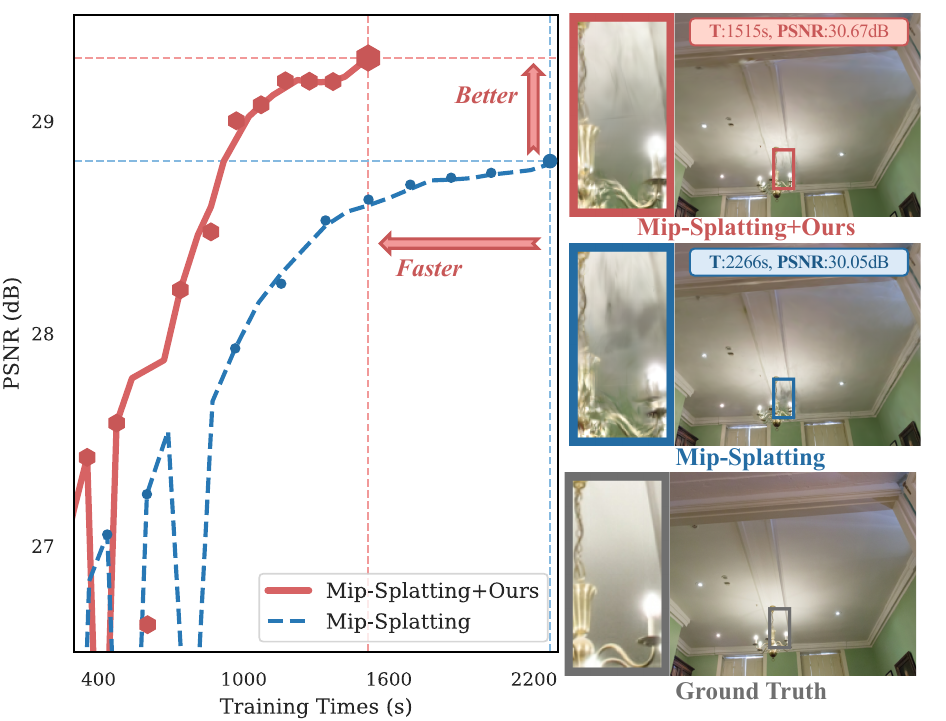}
        \caption{Reconstruction of the “Dr. Johnson”\cite{hedman-2018-deepblending} scene by Mip-Splatting\cite{Yu2024MipSplatting}.}
    \end{subfigure}
    }
    \caption{\textbf{Improvements and Illustrations of Applying Group Training in 3DGS\cite{kerbl3Dgaussians} and Mip-Splatting\cite{Yu2024MipSplatting}.} 
    Grouping Training (solid line) achieves a significant increase in reconstruction speed and superior scene quality compared to baseline methods (dashed line). 
    In the “Stump”\cite{barron2022mipnerf360} scene, Grouping Training delivers a 28\% faster reconstruction while rendering leaf details with greater clarity. In the “Dr. Johnson”\cite{hedman-2018-deepblending} scene, Mip-Splatting with Group Training achieves a 33\% reduction in reconstruction time and effectively suppresses the generation of floating artifacts.
    }
    \label{fig:teaser_1}
\end{figure}

%% file: sec/1_Intro.tex
\section{Introduction}

\label{sec:intro}
%-------------------------------------------------------------------------
Novel view synthesis (NVS) is a pivotal technology in various domains, such as virtual reality~\cite{jiang2024vr}, augmented reality~\cite{zhai2024splatloc}, and autonomous driving~\cite{zhou2024drivinggaussian}. 
Recently, 3D Gaussian Splatting (3DGS)~\cite{kerbl3Dgaussians} achieved significant success in  NVS due to its real-time~\cite{Li_2024_CVPR, peng2024rtgslam,zhu2025fsgs}, high-quality rendering~\cite{bulo2024revising,meng2024mirror}. 
This high-quality rendering is made possible by using millions of Gaussians~\cite{chen2024survey}, each includes a complex set of attributes, namely, position, size, orientation, opacity, and color. 
These attributes are optimized via multi-view photometric losses. \\

%-------------------------------------------------------------------------
% $\triangleright$
\noindent $\bullet$ \textbf{Motivation.} However,  3DGS also introduces a potential risk during training, as the exponentially growing Gaussians significantly increase the training burden. 
Consequently, a primary research focus is the development of methods that enable rapid and cost-effective 3D scene reconstruction without compromising the rendering performance.

During the densification stage, the cyclic pruning of Gaussians with low opacity emerges as a straightforward yet potentially effective strategy~\cite{kerbl3Dgaussians,papantonakisReduced3DGS} for enhancing the training efficiency of 3DGS. Specifically, in each round, 3DGS prunes Gaussians whose opacity falls below a designated threshold $\epsilon_\alpha$, which is a vital parameter that facilitates control over the number of generated Gaussians. However, this method has inherent limitations. 
If $\epsilon_\alpha$ is set too conservatively, the resultant acceleration will be minimal while an excessively lenient threshold defeats the quality of the NVS. 
\textit{Thus, achieving a \textbf{balance} between accelerating the training and preserving the rendering quality poses a considerable challenge.} \\

%-------------------------------------------------------------------------
\noindent $\bullet$ \textbf{Intuition and contributions.} Inspired by the limitations of the cyclic pruning strategy, we propose to \textbf{\textit{cache a portion of the Gaussians}} instead of directly pruning Gaussians. 
This “\textbf{\textit{cache}}” concept can retain the “important” Gaussians while reducing the total number of them.
Therefore, we propose a novel \textbf{\textit{Group Training}} strategy to accelerate the training of 3DGS. 
Specifically, \textit{(1)} a portion of Gaussians is cyclically cached during training.
These cached Gaussians are temporarily excluded from scene rendering and model training, which greatly reduces the number of Under-training Gaussians and cuts down on training time.
\textit{(2)} Additionally, through cyclic resampling, the cached Gaussians are reintegrated into the 3DGS model, reducing the impact of direct pruning.
\textit{(3)} A crucial element of Group Training is determining how to sample Gaussians to distinguish between under-training and cached Gaussians. 
A naive sampling approach would be random sampling.
Interestingly, we discovered that the distribution of Gaussian opacity values influences both the quantity of Gaussians generated during densification and the rendering speed of the scene. 
Therefore, we introduce an \textbf{\textit{Opacity-based Prioritized Sampling}} method, which effectively reduces the generation of redundant Gaussians and improves the training speed of 3DGS.
%-------------------------------------------------------------------------
Specifically, the main technical contributions of our work are as follows:

\begin{enumerate}
    \item We propose a simple yet highly efficient training framework called Group Training for 3D Gaussian Splatting, which is seamlessly integrated into existing 3DGS frameworks, including both 3DGS and Mip-Splatting.
    \item Our framework significantly improves the efficiency of scene reconstruction and the quality of NVS. 
\end{enumerate}

\input{figures/1_intro}

% Our code will be made \textbf{publicly available} upon acceptance of this paper ( but is \textbf{now} available in the supplementary materials for reviewing).

%% file: figures/1_intro.tex
\begin{figure}[!t]
    \centering
    \includegraphics[trim={0.cm 0 0.cm 0},clip,width=0.495\linewidth]{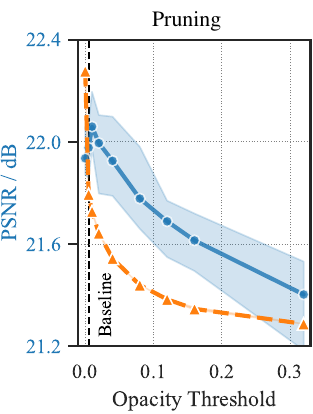}
    \includegraphics[trim={0.cm 0 0.cm 0},clip,width=0.495\linewidth]{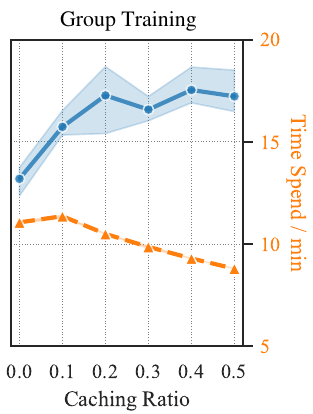}
    \caption{\textbf{PSNR and Time Performance of Pruning~\cite{kerbl3Dgaussians,papantonakisReduced3DGS} and Group Training under Varying Hyperparameters.} The reconstructions are performed on the ``Train'' scene~\cite{knapitsch-2017-tanksandtemples} using 3DGS~\cite{kerbl3Dgaussians} as the baseline. \textbf{Left:} The pruning method exhibits substantial instability in optimizing the trade-off between reconstruction efficiency and quality. The hyperparameter sensitivity (Opacity Threshold) presents significant challenges for optimal parameter tuning. \textbf{Right:} Group Training demonstrates consistent improvements in both reconstruction speed and quality, with robust performance across a wide range of hyperparameter values (Caching Ratio), enabling straightforward parameter optimization.}
    \label{fig:motivationa}
    \vspace{-8pt}
\end{figure}

%% file: sec/2_Related.tex
\input{figures/3_grouping_training}
\section{Related Work}
\label{sec:related_work}
In this section, we review relevant work in novel view synthesis and 3D Gaussian Splatting, followed by the distinction of our approach.

\vspace{5pt}
\noindent $\bullet$ \textbf{NVS.} NVS involves generating new views of a scene from different viewpoints than those in the original dataset.
Neural Radiance Fields (NeRF)~\cite{mildenhall2020nerf} have been pivotal in NVS for producing photo-realistic images through volume rendering. 
Neural Radiance Fields (NeRF) have significantly advanced NVS by producing photo-realistic images via volume rendering, but they require substantial computational resources and lengthy training due to costly MLP evaluations. To overcome these limitations, methods such as voxel grids~\cite{yu2022plenoxels}, hash grids~\cite{mueller2022instant} and points~\cite{xu2022point} have been introduced to facilitate real-time NeRF rendering.

Recently, 3D Gaussian Splatting (3DGS)~\cite{kerbl3Dgaussians} has emerged as a promising alternative in the NVS field.
Unlike the implicit structure of NeRF, 3DGS uses rasterization rather than ray tracing, enabling efficient rendering suitable for applications requiring both speed and image quality. \\

\noindent $\bullet$ \textbf{3DGS.}
3D Gaussian Splatting (3DGS) is a recent approach to high-quality, real-time novel view synthesis that represents scenes explicitly as 3D Gaussian primitives. By rasterizing these Gaussians into 2D splats and blending them using $\alpha$-blending~\cite{mildenhall2020nerf}, 3DGS produces detailed, high-resolution outputs, making it well-suited for applications that demand both speed and visual fidelity.

The 3DGS training process involves two main stages: “Densification” and “Optimization”. During the densification phase (0$\sim$15K iterations), Adaptive Density Control~\cite{kerbl3Dgaussians} is applied to increase the density of Gaussian primitives, thereby enhancing the model’s capacity for detailed scene representation. In the optimization phase (15K$\sim$30K iterations), densification stops, and the existing Gaussians undergo global refinement, yielding a more precise scene depiction.
Recent studies have driven significant advancements in 3DGS across several areas, notably improving rendering quality~\cite{bulo2024revising, Yu2024MipSplatting, kheradmand20243d}, memory efficiency~\cite{niedermayr2023compressed, fan2023lightgaussian, papantonakis2024reducing}, and scalability for large scenes~\cite{liu2025citygaussian}.
Beyond these enhancements, additional research has focused on increasing the efficiency of 3DGS optimizers~\cite{hoellein_2024_3dgslm} and refining Gaussian Densification~\cite{taming3dgs,scaffoldgs, yang2024localized} and Pruning strategies~\cite{fan2023lightgaussian,fang2024minisplattingrepresentingscenesconstrained,niemeyer2024radsplat}. These improvements yield more compact 3DGS models that support faster reconstruction but sacrificing detail.\\

\noindent $\bullet$ \textbf{Distinction.} Unlike previous optimization-focused approaches, we propose Group Training, a principled strategy that dynamically manages Gaussian primitives during training through controlled group sampling. This approach not only accelerates training but also improves reconstruction quality by maintaining effective primitive optimization, achieving a better balance between computational efficiency and visual fidelity.

%% file: figures/3_grouping_training.tex
\begin{figure*}[t]
    \centering
     \begin{subfigure}{0.93\linewidth}    
    \hfill
    \includegraphics[trim={0.cm 0 0.cm 0},clip,width=1.0\linewidth]{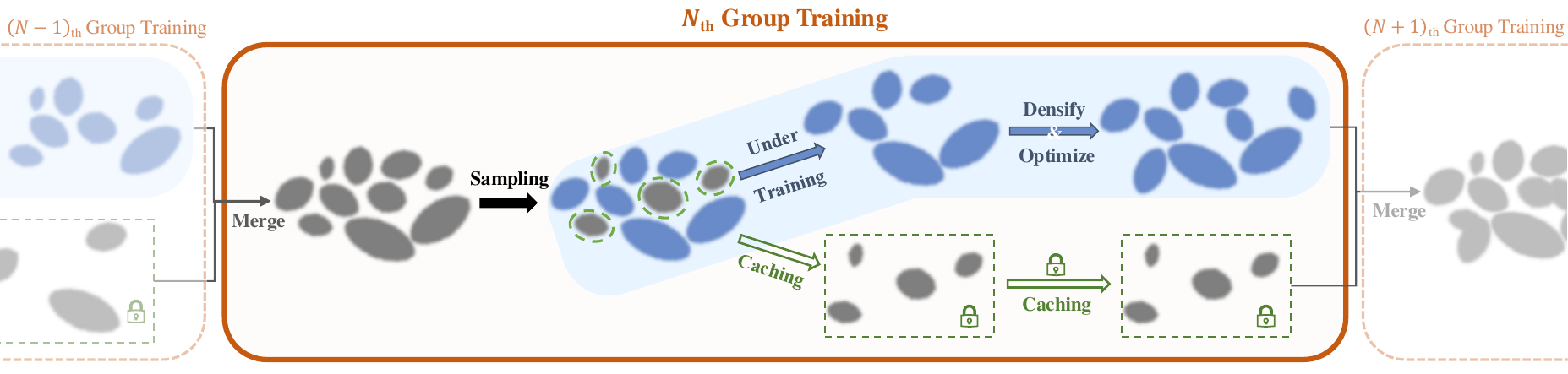}
    \end{subfigure}    
    \caption{\textbf{The overall framework of Group Training. }
    Group Training involves periodically dividing all Gaussian primitives. 
    Specifically, at regular iteration intervals, Gaussians from all groups are merged before rendering the training view. Subsequently, all Gaussian primitives are categorized into the Under-training Group and the Caching Group according to a specified sampling strategy. Before the next grouping, the Under-training Group is utilized for Gaussian densification (Iteration 0$\sim$15K) or optimization (Iteration 15$\sim$30K), while the Caching Group remains inactive and does not participate in any calculations.}
    \label{fig:group_training}
    \vspace{-8pt}
\end{figure*}

%% file: sec/3_Method.tex
\vspace{-5pt}
\section{Methodology}
\label{sec:method}

In~\cref{sec:mtd_Group_Training}, we will provide a comprehensive introduction to the Group Training method and its multifunctional advantages in 3D Gaussian Splatting (3DGS).
% examine its diverse functions across training 3D Gaussian Splatting.
Section~\ref{sec:mtd_Sampling_Strategies} introduces two sampling paradigms: Random Sampling and Prioritized Sampling (incorporating importance score, opacity-based, and volume-based criteria), 
with comparative analysis of their computational efficiency and performance trade-offs. 
We mathematically demonstrate that Opacity-based Prioritized Sampling achieves effective 3DGS densification and efficient rendering.
% Subsequently, we introduce two different sampling strategies: Random Sampling and Prioritized Sampling (including important score, opacity-based, volume-based.) in~\cref{sec:mtd_Sampling_Strategies}, 
% and we compare the computational efficiency and performance loss of various Prioritized Sampling.
% The impact of different Prioritized Sampling strategies on gaussian densification and rendering is also examined.
Finally, in~\cref{sec:mtd_analysis} , we summarize training characteristics of 3DGS with Group Training, and the Sampling Strategies are identified. 

\input{figures/3_schedule_grouping}
\subsection{Group Training}
\label{sec:mtd_Group_Training}
The number of Gaussian primitives in the Gaussian set increases rapidly during the densification phase to achieve rich representation capabilities. 
This implies that the computational time to render a scene from a specific viewpoint will also increase, and this effect will extend to the global optimisation phase.
To this end, we employ the Group Training method during training. 

For any given set of Gaussian~$G$, the Under-training Group~$G_\text{Under-training}$ is obtained through sampling, while the remaining primitives constitute the Cached Group~$G_\text{Cached}$.
{\small
\begin{equation}
    G_{\text{Under-training}} = \left\{ g_i | g_i \in G, i \in I, I \subseteq \{1, 2, 3, ..., |G|\} \right\}
    \label{eq:G_UT}
    \vspace{-12pt}
\end{equation}}
\begin{equation}
    G_{\text {Cached }}=G \backslash G_{\text {Under-training }}
    \label{eq:G_Cached}
\end{equation}
Before the next grouping, the Cached Group is merged with the Under-training Group, after which the new $G_\text{Under-training}$ and $G_\text{Cached}$ groups are resampled. 
A visual representation of the Group Training process is provided in~\cref{fig:group_training}. This procedure ensures that each Gaussian primitive has an opportunity to contribute to the training process, and effectively reducing the number of Under-training Gaussians while retaining those deemed “important.”

As outlined in~\cref{sec:related_work}, the 3DGS training process consists of two distinct phases: “Densification” and “Optimization”. 
Group Training is facilitated in both phases and concludes with Groups Merging at each phases, as illstrated in~\cref{fig:schedule_grouping}.

\subsection{Effective and Efficient Sampling Strategies}
\label{sec:mtd_Sampling_Strategies}

Within each grouping, a subset of Gaussians is sampled from the merged Gaussians. 
The simplest sampling strategy is Random Sampling (RS).
% which selects Gaussians uniformly. 
We propose attribute-driven Prioritized Sampling to optimize 3DGS further. 

We evaluated various attributes, including opacity and volume-based approaches~\cite{fan2023lightgaussian} and importance scores~\cite{niemeyer2024radsplat, fang2024minisplattingrepresentingscenesconstrained, zhang2024lp,girish2023eagles} (see appendix for detailed comparisons). 
Among these, Opacity-based Prioritized Sampling demonstrated superior performance in both training speed and rendering quality.
As mathematically proven in ~\cref{sec:opacity_densify} and~\ref{sec:opacity_render}, opacity intrinsically governs two critical 3DGS properties, both experimentally validated.

% ------------------------------------------------------------
\subsubsection{Sampling Strategy for Better Densification}
\label{sec:opacity_densify}

For any Gaussian primitive $G_m$ projected onto the imaging plane with 2D center coordinates~$[x_\mathrm{m},y_\mathrm{m}]$, the partial derivatives of loss $L$ w.r.t the projected coordinates are 
$[\frac{\partial L}{\partial x_\mathrm{m}},\frac{\partial L}{\partial y_\mathrm{m}}]$, 
which govern the densification criterion by:
\begin{equation}
\vspace{-5pt}
\resizebox{0.5\columnwidth}{!}{$\displaystyle
    \sqrt{ \left( \frac{\partial L}{\partial x_\mathrm{m}} \right)^2 + \left( \frac{\partial L}{\partial y_\mathrm{m}} \right)^2 } > \tau_\mathrm{grad},
$}
\end{equation}
where $\tau_\mathrm{grad}$ denotes the predefined gradient threshold.

We formulate the first fundamental property of 3DGS:

% ============= START: 1 ==============
\noindent $\bullet$ \textbf{Proposition~1.} \textit{\textbf{Opacity-based Effective Gaussians Densification:}}
% During scene reconstruction, 
Gaussian primitives with higher opacity serve as the primary contributors to densification of 3DGS.

\noindent \textit{Proof.}
Under our assumptions of mutual independence between Gaussian attributes and within-primitive parameter independence, the partial derivatives are computed as:
\begin{equation}
\resizebox{0.6\columnwidth}{!}{$\displaystyle
\begin{aligned}
    \frac{\partial L}{\partial x_\mathrm{m}}
        % &= 
        % \sum\limits_\mathrm{pixel}
        % \frac{\partial L}{\partial G^\mathrm{2D}_m}
        % \frac{\partial G^\mathrm{2D}_m}{\partial \Delta x}
        % \frac{\partial \Delta x}{\partial x_\mathrm{m}}    \\
        % &= 
        % \sum\limits_\mathrm{pixel}
        % o_m\frac{\partial L}{\partial \alpha_m}
        % \frac{\partial G^\mathrm{2D}_m}{\partial \Delta x}
        % \frac{\partial \Delta x}{\partial x_\mathrm{m}}    \\
        &= 
        o_m
        \sum\limits_\mathrm{pixel}
        \frac{\partial L}{\partial \hat C}
        \frac{\partial \hat C}{\partial \alpha_m}
        \frac{\partial G^\mathrm{2D}_m}{\partial \Delta x}
        \frac{\partial \Delta x}{\partial x_\mathrm{m}},
        \label{eq:dL_dx}
\end{aligned}
$}
\end{equation}
\begin{equation}
\resizebox{0.6\columnwidth}{!}{$\displaystyle  % 宽度缩放95%
\begin{aligned}
    \frac{\partial L}{\partial y_\mathrm{m}} &= 
        o_m
        \sum\limits_\mathrm{pixel}
        \frac{\partial L}{\partial \hat C}
        \frac{\partial \hat C}{\partial \alpha_m}
        \frac{\partial G^\mathrm{2D}_m}{\partial \Delta y}
        \frac{\partial \Delta y}{\partial y_\mathrm{m}},
        \label{eq:dL_dy}
\end{aligned}
$}
\end{equation}
where $o_m$ denotes the opacity of $G_m$, $\hat{C}$ represents the rendered pixel value, and $\Delta x, \Delta y$ indicate coordinate offsets between pixels and~$[x_\mathrm{m},y_\mathrm{m}]$. 
Under the parameter mutual independence assumptions, the $[\frac{\partial \Delta x}{\partial x_\mathrm{m}},\frac{\partial \Delta y}{\partial y_\mathrm{m}}]$, 
$[\frac{\partial G^\mathrm{2D}_m}{\partial \Delta x}, \frac{\partial G^\mathrm{2D}_m}{\partial \Delta y}]$ 
and $\frac{\partial L}{\partial \hat{C}}$ remain independent of $o_m$. 
% \begin{equation}
%     \mathbb{E} \left[ \frac{\partial L}{\partial \alpha_m} \right] 
%     =\mathbb{E} \left[ \frac{\partial L}{\partial \hat C} \right] \cdot \mathbb{E} \left[ \frac{\partial \hat{C}}{\partial \alpha_m} \right] 
% \end{equation}
We prove that $\frac{\partial \hat{C}}{\partial \alpha_m}$ increases with higher expected value of $o_m$ in appendix:
\begin{equation}
\resizebox{0.6\columnwidth}{!}{$\displaystyle  % 宽度缩放95%
    \mathbb{E} \left[\frac{\partial \hat{C}}{\partial \alpha_m} \right] 
    % = \frac{(c_0-c_\mathrm{bg}) T_{\mathrm{saturation}} }{1-\alpha_0}
    = \frac{(c_0-c_\mathrm{bg}) T_{\mathrm{saturation}} }{1-\mathbb{E}[o_m]  \cdot \mathbb{E}[G_m^\mathrm{2D}]},
    \label{eq:E_dC_dalpha}
$}
\end{equation}
where $c_0$ and $\alpha_0$ denote the expected color and opacity respectively, and $T_{\mathrm{saturation}}$ controls $\alpha$ saturation~\cite{kerbl3Dgaussians}. 

With fixed  $L$,~\cref{eq:dL_dx,eq:dL_dy,eq:E_dC_dalpha} demonstrate that elevated opacity distributions yield greater gradient magnitudes in $\left[\frac{\partial L}{\partial x_\mathrm{m}}, \frac{\partial L}{\partial y_\mathrm{m}}\right]$, thereby making such Gaussians statistically prioritized for densification. \hfill $\Box$

\input{figures/3_densification}
\noindent $\bullet$ \textbf{Experimental Validation: }
Our experiments track Gaussian primitives contributing to densification, focusing on opacity and volumes values. Statistical analysis of attribute distributions during densification steps (Fig.~\ref{fig:densification}) reveals that Gaussians with high opacity and small volume are the primary sources implementing Gaussian densification. 

Notably, insufficient high-opacity Gaussians increase the photometric loss, exacerbating both under-reconstruction and over-reconstruction~\cite{kerbl3Dgaussians}, thereby forces redundant densification.
While small volumes can be employed as a compression metric~\cite{fan2023lightgaussian} in the model compression of 3D Gaussian Splatting, they simultaneously drive reasonable densification.
Take into account all factors, it is advisable to cache Gaussians with low-opacity for better gaussian densification.

% ============= END: 1 ==============

\subsubsection{Sampling Strategy for Faster Rendering}
\label{sec:opacity_render}

During viewpoint-specific image synthesis in 3DGS, the framework implements $\alpha$-blending~\cite{mildenhall2020nerf} through recursive application of the compositing equation on a per-pixel.
\begin{equation}
\resizebox{0.65\columnwidth}{!}{$\displaystyle  % 宽度缩放95%
    C_p=\sum_{i=1}^N c_i \alpha_i T_i, 
    \quad 
    T_i=\prod_{j=1}^{i-1}\left(1-\alpha_j\right), 
    \label{eq:C_p}
$}
\end{equation}
\begin{equation}
\resizebox{0.85\columnwidth}{!}{$\displaystyle  % 宽度缩放95%
    N = \min \left\{
        i \in \mathbb{N}^+ ,\bigg| \prod_{j=1}^{i-1} (1 - \alpha_j) \leq T_{\mathrm{saturation}}
        \right\}
    \label{eq:N}
$}
\end{equation}
% ============================================
As established in 3DGS~\cite{kerbl3Dgaussians}, the determination of $N$ hinges on achieving opacity saturation within a pixel's accumulated $\alpha$ value (Transmittance $T_i < T_{\mathrm{saturation}}$).
This critical threshold implementation introduces a fundamental constraint on rendering efficiency: during forward rendering operations, the sequential traversal of Gaussian primitives inherently prevents parallel computation of $T_i$ values. 
Consequently, $N$ emerges as the principal determinant of rendering speed, where any reduction on $N$ directly translates to accelerated renderings.

% ============= START: 2 ==============

Based on the preceding analysis, we formulate the second fundamental property of the 3DGS framework:

\noindent $\bullet$ \textbf{Proposition~2.} \textit{\textbf{Opacity-based Efficient Rendering Acceleration:}}
Gaussian primitives with higher opacity enable faster rendering through faster achieving $\alpha$ saturation.

\noindent \textit{Proof.}
Building upon the premise of mutually independent $\alpha_i$ values between Gaussians, we posit that the mathematical expectation $\mathbb{E}[T_{N}] \approx T_{\mathrm{saturation}}$, 
\begin{equation}
    \mathbb{E}[T_{N}] 
    % = \mathbb{E}\left[\prod_{i=1}^{N}\left(1-\alpha_i\right)\right]
    = (1-\mathbb{E}\left[{\alpha_i}\right])^{N} 
    = (1-\mathbb{E}[o_i] \cdot \mathbb{E}[G_i^\mathrm{2D}])^N, 
    \label{eq:E_T}
\end{equation}
where $\alpha_i$ combines its intrinsic opacity $o_i$ and $G_i^\mathrm{2D}$ (dependent on view projection and covariance, independent of $o_i$). 
We decouple the expected value of $\alpha_i$. \hfill $\Box$

\noindent $\bullet$ \textbf{Experimental Validation: }We generated a 3DGS model with Gaussian $o_i$ values sampled from $\mathcal{N}(\mu_o, 0.1)$, fixing the other parameters.
The rendering time as a function of $\mu_o$ is shown in \cref{fig:opacity_render}.
% This contrasts with methods like that reduce $o_i$ via regularization, but unaware of its rendering speed implications.
The results reveals that Gaussian primitives with higher-opacity induce faster transmittance saturation, thereby reducing the blending steps $N$ and rendering computational load.

% =============== END: 2 ================

\subsection{Analysis and Design Insights}
\label{sec:mtd_analysis}

Through a comprehensive testing program and a detailed examination of the densification and rendering formulas in 3DGS, 
we have identified the acceleration effect of the opacity attribute in the 3DGS training process. 
Based on these insights, we propose an Opacity-based Prioritised Sampling (OPS) strategy. 

The Opacity-based sampling probability $p_i$ for selecting an Under-training Gaussian $G_i$ is defined as:
\vspace{-5pt}
\begin{equation}
    p_i=\frac{\alpha_i}{\sum_{i=1}^N \alpha_i },
    \label{eq:p_under_training}
\end{equation}
where $\alpha_i$ represents the opacity of Gaussian primitive $G_i$, and $N$ is the total number of Gaussian primitives.

\input{figures/3_opacity_render}

%% file: figures/3_schedule_grouping.tex
\begin{figure}[!t]
    \centering
    \includegraphics[width=0.95\linewidth]{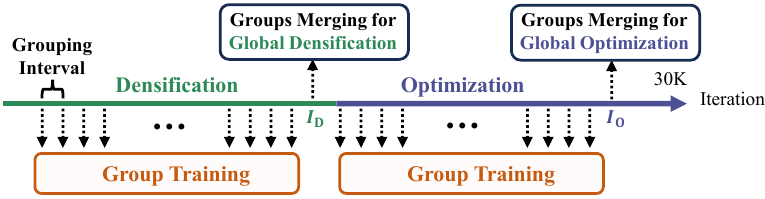}
    \caption{\textbf{Schedule for activating Group Training during reconstruction. }
    Group Training is enabled at regular intervals. \textit{Groups Merging (Grouping without resampling)} occurring at iteration $I_{\text{D}}$ and $I_{\text{O}}$ before the completion of the densification and optimization processes for global densification and optimization.}
    \label{fig:schedule_grouping}
    \vspace{-10pt}
\end{figure}

%% file: figures/3_densification.tex
\begin{figure}[t]
    \centering
    \begin{subfigure}{1\linewidth}
        \centering
        \includegraphics[trim={0.cm 0 0.cm 0}, clip, width=0.49\linewidth]{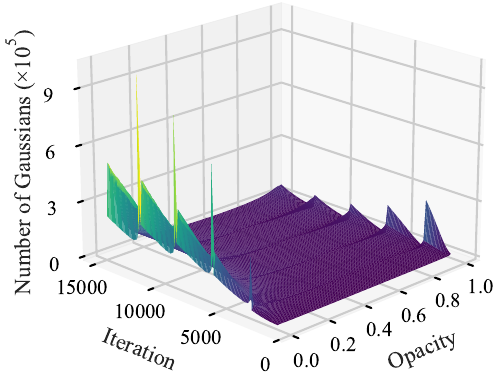}
        \includegraphics[trim={0.cm 0 0.cm 0}, clip, width=0.49\linewidth]{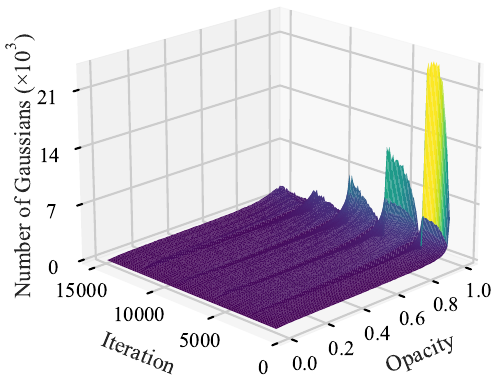}
        \caption*{The distribution of opacity.}
    \end{subfigure}
    \begin{subfigure}{1\linewidth}
        \centering
        \includegraphics[trim={0.cm 0 0.cm 0}, clip, width=0.49\linewidth]{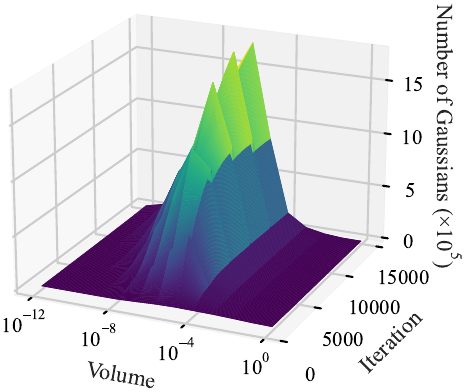}
        \includegraphics[trim={0.cm 0 0.cm 0}, clip, width=0.49\linewidth]{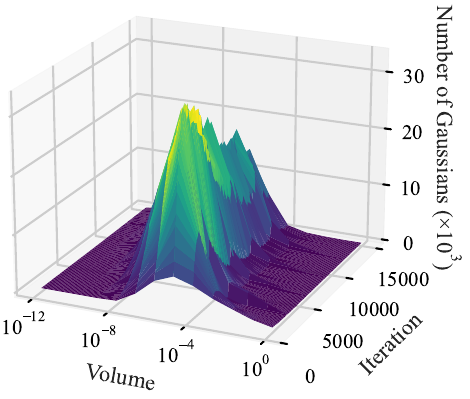}
        \caption*{The distribution of volume.}
    \end{subfigure}
    \caption{
        \textbf{The distribution of Gaussian attributes.}
        The distribution of all Gaussian attributes (left) and those contributing specifically to densification (right) in the “Bicycle”\cite{barron2022mipnerf360} with 3DGS~\cite{kerbl3Dgaussians}.
        \textbf{Top row:} While the opacities are primarily concentrated around 0 and 1, the Gaussians that contribute to densification are predominantly situated around 1.
        \textbf{Bottom row:} The distribution of Volume. As densification progresses, Gaussians with lower volume become increasingly involved in the densification process.
    }
    \vspace{-10pt}
    \label{fig:densification}
\end{figure}

%% file: figures/3_opacity_render.tex
\begin{figure}[t]
    \centering
    \vspace{3pt}
    \resizebox{0.99\linewidth}{!}{
    \includegraphics[width=0.5\linewidth]{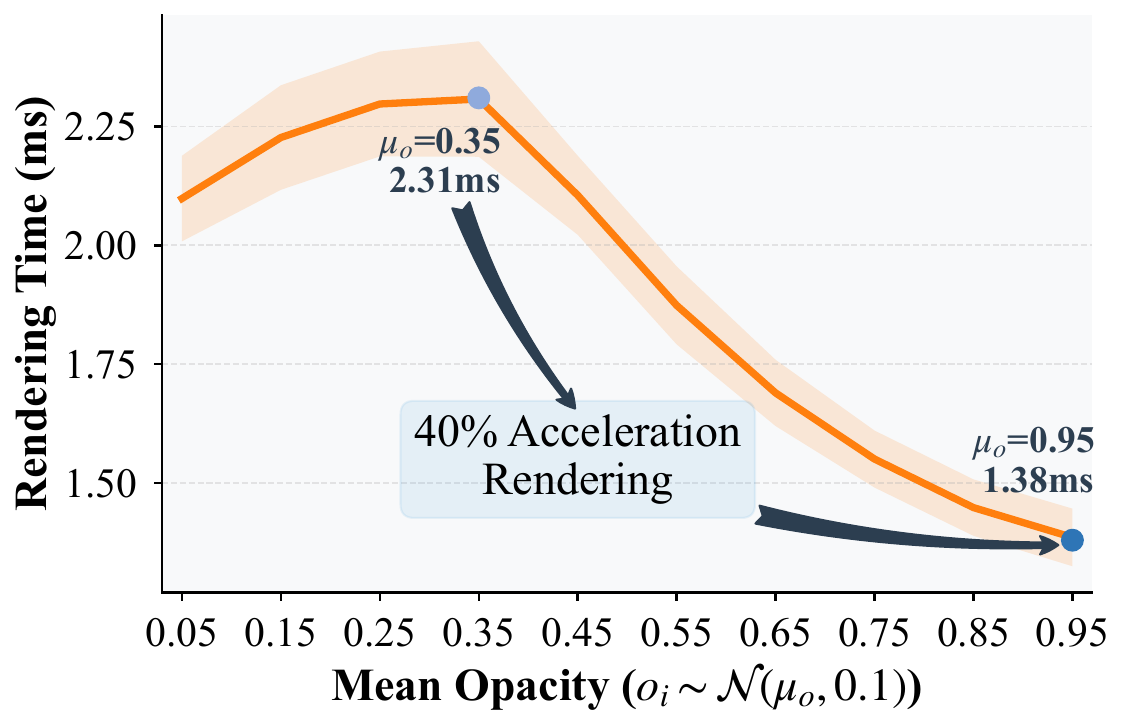}
    \hspace{-5pt}
    \includegraphics[width=0.5\linewidth]{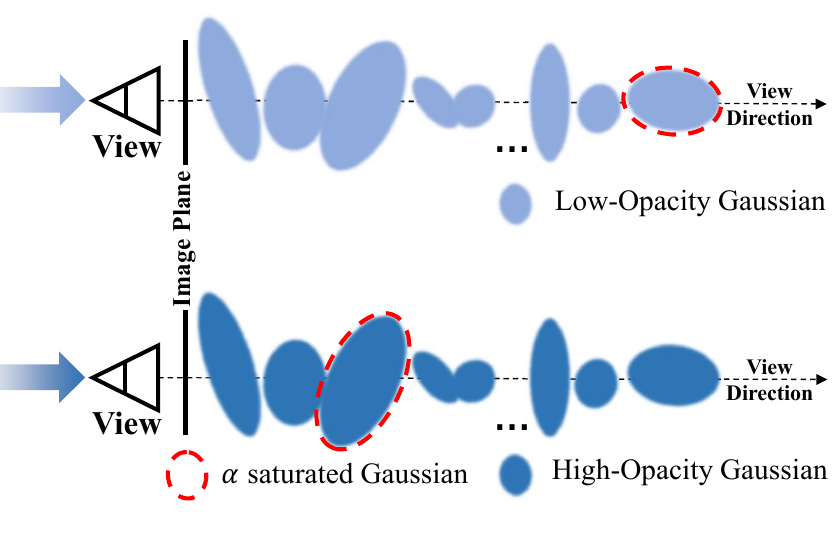}
    }
    % \vspace{-10pt}
    \vspace{-7pt}
    % \caption{\textbf{High-opacity Gaussians accelerate the rendering of 3DGS.} 
    % The opacity distribution significantly impacts 3DGS rendering speed.
    % \textbf{Left:} High-opacity distributions enhance 3DGS rendering speed.
    % \textbf{Right:} The mechanism by which high-opacity Gaussians promote saturation. 
    % }
    \caption{\textbf{High-opacity Gaussians accelerate 3DGS rendering.} 
    \textbf{Left: }The rendering time decreases by 40\% as $\mu_o$ increases. 
    \textbf{Right: }The mechanisms that highers-opacity Gaussians achieve faster $\alpha$ saturation with less number of Gaussians.
    % through concentrated energy deposition along view directions. 
    % The saturation process is mediated by opacity-dependent ray termination probabilities during splatting.
    }
    \label{fig:opacity_render}
    \vspace{-13pt}
\end{figure}

%% file: sec/4_Experiments.tex
\input{tables/4_exp/3dgs_time}
\input{tables/4_exp/3dgs_quality}

\input{tables/4_exp/mip_s_quality}

\input{tables/4_exp/mip_s_mipNerf360}

\input{figures/4_visualization_comp}
\input{tables/4_exp/ablation_tanks}

% --------------------------------------------------

\section{Experiments}
\label{sec:experiments}

We undertake an experimental investigation of Group Training within the context of 3D Gaussian Splatting~\cite{kerbl3Dgaussians} and Mip-Splatting~\cite{Yu2024MipSplatting}. 
The experimental settings are detailed in~\cref{sec:exp_setting}. 
Subsequently, we present the experimental results in~\cref{sec:exp_results}. 
Cross-baselines applicability validation across diverse scenarios is further presented in~\cref{sec:method_appli}. 
The ablation experiments and their results are comprehensively discussed in~\cref{sec:exp_ablation}. 
Finally, we analyse the results of different sampling strategies in ~\cref{sec:exp_discussion}.

\subsection{Experimental Settings}
\label{sec:exp_setting}
\noindent \textbf{Implementation Details. }
The Group Training method was integrated as a plug-in to facilitate its utilisation in conjunction with 3DGS. 
% Specifically, Group Training was incorporated before rendering each training iteration, with its activation controlled by the iteration index. 
We implemented our method within both 
3D Gaussian Splatting~\cite{kerbl3Dgaussians}\footnote{\url{https://github.com/graphdeco-inria/gaussian-splatting}} and 
Mip-Splatting~\cite{Yu2024MipSplatting}\footnote{\url{https://github.com/autonomousvision/mip-splatting}}.

Group Training was implemented with a default grouping interval of 500 iterations throughout the training process. We employed both Random Sampling (RS) and Opacity-Based Prioritized Sampling (OPS) strategies with an Under-Training Ratio (UTR) of 0.6. To ensure computational stability, considering the importance of initial Gaussians~\cite{chung2023depth}, Group Training was activated at the 500th iteration after densification begins. The global densification ($I_\text{D}$) and optimization ($I_\text{O}$) were scheduled at 14.5K and 29K iterations, respectively.
To manage the increased GPU memory consumption during group merging, we fixed the SH coefficients~\cite{papantonakisReduced3DGS, niedermayr2023compressed} at iterations $I_\text{D}$ and $I_\text{O}$. We also conducted experiments applying Group Training solely during the densification phase. 
% Complete experimental results are presented in~\cref{sec:exp_results}.

\noindent \textbf{Dataset and Metrics. }
The efficacy of our method is evaluated on the 
Mip-NeRF360~\cite{barron2022mipnerf360}\footnote{\url{https://jonbarron.info/mipnerf360/}}, 
Tanks~\&~Temples~\cite{knapitsch-2017-tanksandtemples}\footnote{\url{https://www.tanksandtemples.org/}}, 
Deep Blending~\cite{hedman-2018-deepblending}\footnote{\url{http://visual.cs.ucl.ac.uk/pubs/deepblending/}}, 
and NeRF-Synthetic~\cite{mildenhall2020nerf}\footnote{\url{https://www.matthewtancik.com/nerf}} using a single RTX 3090. 
Subsequently, the model's reconstruction performance is evaluated using PSNR, SSIM, and LPIPS. 
Furthermore, the training time, peak memory usage, and model size for each scene are recorded. 

\subsection{Experiments Results}
\label{sec:exp_results}
This section presents the experimental results of integrating the Group-Training method into the 3D Gaussian Splatting~\cite{kerbl3Dgaussians} and Mip-Splatting~\cite{Yu2024MipSplatting}. 
The test outcomes demonstrate that the Group Training with OPS offers a significant reduction in training time compared to the baseline in all scenes, while concurrently enhancing the rendering quality of the novel view, as presented in~\cref{fig:visual-comp}.

% =========================================================
\vspace{5pt}
\noindent \textbf{3D Gaussian Splatting. }
We first tested the effectiveness of Group Training based on 3DGS. 
The reconstruction efficiency for all scenes is summarized in~\cref{tab:3dgs_time}, and~\cref{tab:3dgs_quality} presents the NVS rendering quality across all scenes.

% ------------------------- Group Training -------------------------
Across these scenes, Group Training consistently resulted in faster reconstruction, and this speedup was more pronounced in complex scenes. 
% In addition to accelerating scene reconstruction, Group Training also steadily improved the quality of reconstruction.
% ------------------------------- RS ------------------------------
However, Group Training with RS could lead to an overabundance of redundant Gaussian primitives in some reconstruction tasks, which complicates subsequent operations, such as model compression~\cite{niedermayr2023compressed,navaneet2023compact3d,papantonakisReduced3DGS}. Additionally, GPU memory usage significantly increased due to this redundancy~\cite{lu2024poisonsplat}, as shown in~\cref{tab:3dgs_time}.
% ------------------------------- OPS ------------------------------
In comparison to Group Training with RS, Group Training with OPS demonstrated a more pronounced improvement in reconstruction speed. For the reconstruction of all three complex scenes, the average speed increase was about 30\%. 
% and for the reconstruction of simpler scenes, the speed increase was about 20\%. 
Moreover, OPS greatly reduced the generation of redundant Gaussian primitives, leading to a smaller reconstruction model size than the original 3DGS reconstruction model in all tasks. Across all reconstruction tasks, the model size was reduced by between 10\% and 40\%, resulting in a more compact model overall.

\noindent \textbf{Mip-Splatting. }
%-------------------------------------------------------------------------
We tested Mip-Splatting with Group Training, another significant variant of the Gaussian Splatting model. 
The experimental results are presented in~\cref{tab:mip_s_quality} and~\cref{tab:mip_s_mipnerf360}. Overall, the performance of Group Training with Mip-Splatting was similar to
% that observed with 
3DGS.
% In our experiments, 
Group Training with RS consistently improved both reconstruction quality and speed for Mip-Splatting but produced the highest number of Gaussian primitives during the densification phase. 
Group Training with OPS achieved the fastest reconstruction speed, and its reconstruction quality was nearly optimal. 
The OPS reduced the proportion of densification and resulted in a smaller model size that was either the smallest among the tested methods or comparable to the baseline.

\vspace{10pt}
\noindent \textbf{Counterintuitive Findings. }
Our experiments reveal two notable findings that challenge conventional understanding:
\begin{enumerate}
    \item Despite producing larger model sizes, Group Training with RS consistently improves reconstruction speed, indicating that training dynamics rather than model size dominates reconstruction efficiency.
    \item Group Training with OPS simultaneously achieves better reconstruction quality with reduced model size. This demonstrates that superior performance stem from its fundamental impact on the optimization process rather than from model size.
\end{enumerate}

%-------------------------------------------------------------------------
\subsection{Methodological Applicability}
\label{sec:method_appli}

Group-Training framework fundamentally addresses accelerated training and rendering in 3DGS through its dataset-agnostic improvement. 
Designed as a pluggable training component, the method demonstrates dual efficacy in both 3DGS acceleration model~\cite{fang2024minisplattingrepresentingscenesconstrained} and compression model~\cite{fan2023lightgaussian}, which is detailed in appendix.

%-------------------------------------------------------------------------
\subsection{Ablation Studies}
\label{sec:exp_ablation}

We conducted an ablation study on the Tanks~\&~Temples dataset~\cite{knapitsch-2017-tanksandtemples} to evaluate the impact of three key components of the Group Training method. The detailed comparison results are provided in~\cref{tab:ablation_tanks}.

The results indicate that cyclic resampling contributes the most to improving training speed. 
Specifically, during iterations $I_\text{D}$$\sim$15K, the Global Densification process adaptively performs cyclic pruning of low-opacity Gaussians within 3DGS, reducing redundant Gaussians, decreasing the model size, and thereby enhancing training efficiency.
When Global Densification is enabled, the 3DGS model removes most redundant Gaussians during global densification, resulting in fewer redundant Gaussians in the Caching Group during the Optimization phase. In this scenario, Global Optimization can make full use of all Gaussians, further improving reconstruction performance.

% -------------------------------------------------------
\vspace{-2pt}
\subsection{Discussion}
\label{sec:exp_discussion}
% ------------------------- A: Why Faster --------------------------
Our experimental results demonstrate that enabling Group Training and utilizing OPS improved both reconstruction speed and quality. To better clarify these improvements, two key questions arise from these findings: 

% Group Training
\noindent $\bullet$ \textbf{Why faster?} Group Training significantly accelerates reconstruction primarily by reducing the number of Under-training Gaussian primitives, thereby decreasing the rendering burden on the model. This speedup is observed even when the final model size is larger than the baseline.
% OPS
In alignment with our earlier analysis in~\cref{sec:mtd_Sampling_Strategies}, the OPS also contributes notably to this speedup effect. 
During the densification phase, Group Training with OPS preserves high-opacity Gaussians within the Under-Training Group, without which could lead to instances of both under-reconstruction and over-reconstruction~\cite{kerbl3Dgaussians,ye2024absgs}. 
These issues can lead to redundant densification. 
As a result, models reconstructed with Group Training and OPS contain fewer redundant Gaussians, leading to smaller model sizes and faster reconstruction overall.
% rendering
Additionally, OPS prioritizes sampling high-opacity Gaussians, which rapidly drives $\alpha$ saturation, which also discussed in~\cref{sec:mtd_Sampling_Strategies}. 
Focusing on these most impactful Gaussians reduces the number of Gaussians needed for splatting, further enhancing training and rendering efficiency.

% ------------------------- A: Why Better --------------------------
\noindent $\bullet$ \textbf{Why better?}
% Group Training with OPS prioritizes the retention and optimization of high-opacity Gaussian primitives, ensuring that elements most crucial to scene rendering are refined more effectively. the similar fidelity with the smaller model than RS, or even smaller than baseline in many cases. 
Group Training with the OPS prioritizes the retention and optimization of high-opacity Gaussian primitives. 
This achieves comparable or superior fidelity to RS, while significantly reducing the model size.
The Gaussian primitives with higher opacity inherently have a greater influence on the final rendered image~\cite{fan2023lightgaussian, papantonakisReduced3DGS,taming3dgs,morgenstern2023compact}, so emphasizing their optimization leads to significant improvements in overall rendering quality.
Besides, cyclic resampling temporarily excluding a randomly selected subset of Gaussian primitives, each remaining primitive is compelled to learn how to independently represent essential scene features leading to a more robust representation.

\textbf{In summary,} by integrating these strategies, Group Training with OPS simultaneously enhances both reconstruction speed and rendering quality. By focusing on the most impactful primitives and encouraging independent feature representation, Group Training achieves efficient, high-quality scene reconstruction.

%% file: tables/4_exp/3dgs_time.tex
\begin{table*}[t]
  \centering
  \resizebox{\textwidth}{!}{ % \footnotesize % \small
  \begin{tabular}{@{}ccccccccccccccc@{}}
    \toprule
      ~ & \multirow{2}{*}[-0.8ex]{\shortstack{Grouping\\Iterations}} 
      & \multicolumn{3}{c}{Mip-NeRF360~\cite{barron2022mipnerf360}} 
      & \multicolumn{3}{c}{Tanks~\&~Temples~\cite{knapitsch-2017-tanksandtemples}}  
      & \multicolumn{3}{c}{Deep Blending~\cite{hedman-2018-deepblending}} 
      & \multicolumn{3}{c}{Blender~\cite{mildenhall2020nerf}} 
      \\
     \cmidrule(r){3-5} \cmidrule(r){6-8} \cmidrule(r){9-11} \cmidrule(r){12-14}
      ~ & ~ & PM $\downarrow$ & Size $\downarrow$ & Time $\downarrow$ &
              PM $\downarrow$ & Size $\downarrow$ & Time $\downarrow$ &
              PM $\downarrow$ & Size $\downarrow$ & Time $\downarrow$ &
              PM $\downarrow$ & Size $\downarrow$ & Time $\downarrow$
       \\
     % ---------------------------------------------------
    \midrule
        3D Gaussian Splatting$*$  
        ~ & -- & \cellcolor{orange!25}8.69  & 792  & 26.72  & 4.59  & \cellcolor{orange!25}434  & 15.00  & 7.72  & 666  & 23.90  & 2.78  & 69  & 6.06   \\ 
        \midrule
        \multirow{2}{*}{Group Training w/ RS} 
        ~ & 0$\sim$15K & 9.51  & 907  & 26.62  & 5.08  & 496  & 14.60  & 7.64  & 644  & 21.85  & 2.76  & \cellcolor{orange!25}63  & 5.66   \\
        ~ & 0$\sim$30K & 9.48  & 902  & 22.58  & 5.04  & 495  & \cellcolor{orange!25}12.25  & 7.63  & \cellcolor{orange!25}643  & \cellcolor{orange!25}18.45  & 2.76  & \cellcolor{orange!25}63  & 5.26   \\ 
        \midrule
        \multirow{2}{*}{Group Training w/ OPS}
        ~ & 0$\sim$15K & \cellcolor{red!25}\textbf{8.57}  & \cellcolor{red!25}\textbf{678}  & \cellcolor{orange!25}22.53  & \cellcolor{red!25}\textbf{4.54}  & \cellcolor{red!25}\textbf{384}  & 12.80  & \cellcolor{orange!25}6.81  & \cellcolor{red!25}\textbf{487}  & 19.25  & \cellcolor{orange!25}2.65  & \cellcolor{red!25}\textbf{43}  & \cellcolor{orange!25}5.01   \\ 
        ~ & 0$\sim$30K & \cellcolor{red!25}\textbf{8.57}  & \cellcolor{orange!25}679  & \cellcolor{red!25}\textbf{19.56}  & \cellcolor{orange!25}4.55  & \cellcolor{red!25}\textbf{384}  & \cellcolor{red!25}\textbf{10.95}  & \cellcolor{red!25}\textbf{6.80}  & \cellcolor{red!25}\textbf{487}  & \cellcolor{red!25}\textbf{16.85}  & \cellcolor{red!25}\textbf{2.64}  & \cellcolor{red!25}\textbf{43}  & \cellcolor{red!25}\textbf{4.79}   \\
    \bottomrule
     % ---------------------------------------------------
  \end{tabular}
  }
    \caption{\textbf{Comparison of reconstruction efficiency for 3DGS~\cite{kerbl3Dgaussians}. }
  Group Training significantly enhances reconstruction speed across all four datasets, with the OPS strategy achieving the highest acceleration. This effect is particularly notable in complex scenes. While Group Training with RS produces a more redundant model, the OPS strategy yields more compact models and reduced GPU peak memory usage.
  $*$ indicates that the model was retrained. 
  PM stands for GPU peak memory allocation, with Size in MB and Time in minutes.}
  \label{tab:3dgs_time}
\end{table*}

%% file: tables/4_exp/3dgs_quality.tex
\begin{table*}[ht]
  \centering
  \resizebox{\textwidth}{!}{ % \footnotesize % \small
  \begin{tabular}{@{}ccccccccccccccccccc@{}}
    % \centering
    \toprule
      ~ & \multirow{2}{*}[-0.8ex]{\shortstack{Grouping\\Iterations}} 
        & \multicolumn{4}{c}{Mip-NeRF360~\cite{barron2022mipnerf360}} 
        & \multicolumn{4}{c}{Tanks\&Temples~\cite{knapitsch-2017-tanksandtemples}} 
        & \multicolumn{4}{c}{Deep Blending~\cite{hedman-2018-deepblending}}  
        & \multicolumn{4}{c}{Blender~\cite{mildenhall2020nerf}} \\
     \cmidrule(r){3-6} \cmidrule(r){7-10} \cmidrule(r){11-14} \cmidrule(r){15-18}
      ~ & ~ & PSNR $\uparrow$ & SSIM $\uparrow$ & LPIPS $\downarrow$ & Time $\downarrow$ 
            & PSNR $\uparrow$ & SSIM $\uparrow$ & LPIPS $\downarrow$ & Time $\downarrow$ 
            & PSNR $\uparrow$ & SSIM $\uparrow$ & LPIPS $\downarrow$ & Time $\downarrow$ 
            & PSNR $\uparrow$ & SSIM $\uparrow$ & LPIPS $\downarrow$ & Time $\downarrow$ \\
     % ---------------------------------------------------
    \midrule
        3D GS~\cite{kerbl3Dgaussians} & -- 
           & 27.205 & 0.815 & 0.2143 & --
           & 23.142 & 0.841 & 0.183  & --  
           & 29.405 & 0.903  & \cellcolor{red!25}\textbf{0.2425} & --
           & 33.33 & 0.969 & 0.030 & --\\
           % ------------------------------------------
        3D GS$*$ & -- 
        & 27.445  & 0.816  & 0.2155  & 26.7 
          & 23.697  & 0.849  & 0.1764  & 15.0  
          & 29.586  & 0.904  & 0.2437  & 23.9
          & 33.772  & 0.970  & 0.0306  & 6.1 \\ 
        \midrule
        \multirow{2}{*}{\shortstack{Group Training\\w/ RS}}
        ~ & 0$\sim$15K 
        & \cellcolor{red!25}\textbf{27.621}  & \cellcolor{red!25}\textbf{0.822}  & \cellcolor{red!25}\textbf{0.2074}  & 26.6
        & 23.773  & \cellcolor{red!25}\textbf{0.851}  & \cellcolor{red!25}\textbf{0.1703}  & 14.6  
         & 29.185  & 0.901  & 0.2473  & 21.9
         & \cellcolor{red!25}\textbf{33.959}  & \cellcolor{red!25}\textbf{0.971}  & \cellcolor{red!25}\textbf{0.0284}  & 5.7  \\
         % ------------------------------------------
        ~ & 0$\sim$30K  
        & 27.537  & \cellcolor{orange!25}0.821  & 0.2157  & 22.6
        & 23.703  & 0.848  & 0.1823  & \cellcolor{orange!25}12.3  
        & 29.417  & \cellcolor{red!25}\textbf{0.907}  & 0.2481  & \cellcolor{orange!25}18.5 
        & 33.877  & \cellcolor{red!25}\textbf{0.971}  & 0.0295  & 5.3 \\ 
        \midrule
        % \multirow{2}{*}[-0.8ex]{\shortstack{Grouping\\Iterations}} 
        \multirow{2}{*}{\shortstack{Group Training\\w/ OPS}}
        ~ & 0$\sim$15K
        & \cellcolor{orange!25}27.582  & 0.820  & \cellcolor{orange!25}0.2103  & \cellcolor{orange!25}22.5 
        & \cellcolor{orange!25}23.842  & \cellcolor{red!25}\textbf{0.851}  & \cellcolor{orange!25}0.1726  & 12.8  
        & \cellcolor{orange!25}29.713  & \cellcolor{orange!25}0.906  & \cellcolor{orange!25}0.2435  & 19.3 
        & \cellcolor{orange!25}33.889  & \cellcolor{red!25}\textbf{0.971}  & \cellcolor{orange!25}0.0292  & \cellcolor{orange!25}5.0  \\ 
        % ------------------------------------------
        ~ & 0$\sim$30K
        & 27.564  & 0.820  & 0.2133  & \cellcolor{red!25}\textbf{19.6}
        & \cellcolor{red!25}\textbf{23.853}  & \cellcolor{orange!25}0.850  & 0.1764  & \cellcolor{red!25}\textbf{11.0} 
         & \cellcolor{red!25}\textbf{29.746}  & \cellcolor{red!25}\textbf{0.907}  & 0.2450  & \cellcolor{red!25}\textbf{16.9}   & 33.808  & \cellcolor{orange!25}0.970  & 0.0299  & \cellcolor{red!25}\textbf{4.8}  \\ 
    \bottomrule
     % ---------------------------------------------------
  \end{tabular}
  }
  \caption{\textbf{Comparison of reconstruction quality for 3DGS~\cite{kerbl3Dgaussians}. }
  % Group Training simultaneously accelerates the 3DGS scene reconstruction process and enhances the accuracy of 3D reconstructions. Specifically, Group Training with OPS achieves a notable improvement in reconstruction quality while maintaining the highest reconstruction speed.
  Group Training simultaneously accelerates the 3DGS scene reconstruction process and improves the accuracy of 3D reconstructions. Notably, Group Training with OPS achieves a notable improvement in reconstruction quality while maintaining the highest reconstruction speed.
  % Enabling Group Training during the optimization stage~(15K~$\sim$~30K iteration) has minimal impact on reconstruction quality but yields a substantial increase in reconstruction speed.
  }
  \label{tab:3dgs_quality}
  \vspace{-5pt}
\end{table*}

%% file: tables/4_exp/mip_s_quality.tex
\begin{table*}[ht]
  \centering
  \resizebox{\textwidth}{!}{ % \footnotesize % \small
  \begin{tabular}{@{}ccccccccccccccccc@{}}
    % \centering
    \toprule
      ~ & \multirow{2}{*}[-0.8ex]{\shortstack{Grouping\\Iterations}} 
        & \multicolumn{4}{c}{Tanks~\&~Temples~\cite{knapitsch-2017-tanksandtemples}} 
        & \multicolumn{4}{c}{Deep Blending~\cite{hedman-2018-deepblending}}  
        & \multicolumn{4}{c}{Blender~\cite{mildenhall2020nerf}} \\
     \cmidrule(r){3-6}
     \cmidrule(r){7-10} \cmidrule(r){11-14} 
      ~ & ~ & PSNR $\uparrow$ & SSIM $\uparrow$ & LPIPS $\downarrow$ & Time $\downarrow$ 
            & PSNR $\uparrow$ & SSIM $\uparrow$ & LPIPS $\downarrow$ & Time $\downarrow$ 
            & PSNR $\uparrow$ & SSIM $\uparrow$ & LPIPS $\downarrow$ & Time $\downarrow$ \\
     % ---------------------------------------------------
    \midrule
        Mip-Splatting$*$ & -- 
        & 23.749  & 0.860  & 0.1562  & 23.0
        & 29.358  & 0.903  & 0.2390  & 35.1
        & 33.995  & 0.970  & 0.0296  & 8.5\\ 
        \midrule
        \multirow{2}{*}{\shortstack{Group Training\\w/ RS}}
        ~ & 0$\sim$15K 
        & 23.953  & \cellcolor{red!25}\textbf{0.863}  & \cellcolor{red!25}\textbf{0.1504}  & 24.6
        & 28.929  & 0.900  & 0.2431  & 31.7
        & \cellcolor{red!25}\textbf{34.203}  & \cellcolor{orange!25}0.971  & \cellcolor{red!25}\textbf{0.0278}  & 9.3   \\
         % ------------------------------------------
        ~ & 0$\sim$30K  
        & \cellcolor{orange!25}24.146  & \cellcolor{orange!25}\textbf{0.862}  & 0.1598  & \cellcolor{orange!25}20.4
        & 29.392  & \cellcolor{red!25}\textbf{0.908}  & 0.2420  & \cellcolor{orange!25}26.5
        & 34.123  & \cellcolor{red!25}\textbf{0.972}  & \cellcolor{orange!25}0.0285  & 8.3  \\ 
        \midrule
        \multirow{2}{*}{\shortstack{Group Training\\w/ OPS}}
        ~ & 0$\sim$15K
        & 23.958  & \cellcolor{red!25}\textbf{0.863}  & \cellcolor{orange!25}0.1519  & 21.7
        & \cellcolor{orange!25}29.665  & \cellcolor{orange!25}0.907  & \cellcolor{red!25}\textbf{0.2364}  & 28.1
        & \cellcolor{orange!25}34.132  & \cellcolor{orange!25}0.971  & \cellcolor{orange!25}0.0285  & \cellcolor{orange!25}8.2  \\ 
        % ------------------------------------------
        ~ & 0$\sim$30K
        & \cellcolor{red!25}\textbf{24.156}  & \cellcolor{red!25}\textbf{0.863}  & 0.1559  & \cellcolor{red!25}\textbf{18.2}
        & \cellcolor{red!25}\textbf{29.788}  & \cellcolor{red!25}\textbf{0.908}  & \cellcolor{orange!25}0.2384  & \cellcolor{red!25}24.0
        & 34.110  & \cellcolor{orange!25}0.971  & 0.0286  & \cellcolor{red!25}\textbf{7.5}\\ 
    \bottomrule

  \end{tabular}
  }
  \caption{\textbf{Comparison of reconstruction quality for Mip-Splatting~\cite{Yu2024MipSplatting}. } 
  Group Training achieves simultaneous improvements in both speed and reconstruction quality for Mip-Splatting across all tasks. 
  Specifically, Group Training with OPS attains the fastest reconstruction speed across all three datasets and delivers the highest reconstruction quality in the Tanks~\&~Temples and Deep Blending dataset.
  }

  \label{tab:mip_s_quality}
\end{table*}

%% file: tables/4_exp/mip_s_mipNerf360.tex
\begin{table*}
  \centering
  \resizebox{\textwidth}{!}{
  \begin{tabular}{@{}cccccccccccccccccc@{}}
    \toprule
      ~ & \multirow{2}{*}[-0.8ex]{\shortstack{Grouping\\Iterations}} & \multicolumn{5}{c}{Outdoor} & \multicolumn{5}{c}{Indoor}  & \multicolumn{5}{c}{Average} \\
     \cmidrule(r){3-7} \cmidrule(r){8-12} \cmidrule(r){13-17}
      ~ & ~ & PSNR $\uparrow$ & SSIM $\uparrow$ & LPIPS $\downarrow$ & Size $\downarrow$ & Time $\downarrow$ 
            & PSNR $\uparrow$ & SSIM $\uparrow$ & LPIPS $\downarrow$ & Size $\downarrow$ & Time $\downarrow$ 
            & PSNR $\uparrow$ & SSIM $\uparrow$ & LPIPS $\downarrow$ & Size $\downarrow$ & Time $\downarrow$ \\
    \midrule
        Mip-Splatting$*$ & --  
        & 24.814  & 0.748  & 0.1995 & \cellcolor{red!25}\textbf{1430} & 43.3  
        & 31.265  & 0.929  & 0.1738 & \cellcolor{orange!25}448  & 33.3  
        & 27.681  & 0.828  & 0.1881 & \cellcolor{red!25}\textbf{994}  & 38.8  \\ 
        \midrule
        % ------------------------------------------
        \multirow{2}{*}{\shortstack{Group Training\\w/ RS}}
        ~ & 0$\sim$15K
        & 24.969  & \cellcolor{orange!25}0.757  & \cellcolor{red!25}\textbf{0.1897} & 2017 & 45.9  
        & 31.266  & \cellcolor{red!25}\textbf{0.931}  & \cellcolor{red!25}\textbf{0.1697} & 556 & 34.9  
        & 27.768  & \cellcolor{orange!25}0.834  & \cellcolor{red!25}\textbf{0.1808} & 1377 & 41.0
        \\
         % ------------------------------------------
        ~ & 0$\sim$30K  
        & \cellcolor{red!25}\textbf{25.216}  & \cellcolor{red!25}\textbf{0.763}  & 0.1969 & 2017 & \cellcolor{orange!25}36.8  
        & 31.164  & 0.929  & 0.1757 & 556 & 30.5  
        & \cellcolor{red!25}\textbf{27.860}  & \cellcolor{red!25}\textbf{0.837}  & 0.1875 & 1377 & \cellcolor{orange!25}34.0  \\ 
        \midrule
        % ------------------------------------------
        \multirow{2}{*}{\shortstack{Group Training\\w/ OPS}}
        ~ & 0$\sim$15K
        & 24.981  & 0.754  & \cellcolor{orange!25}0.1940 & \cellcolor{orange!25}1571 & 40.7  
        & \cellcolor{red!25}\textbf{31.417}  & \cellcolor{red!25}\textbf{0.931}  & \cellcolor{orange!25}0.1725  & \cellcolor{red!25}\textbf{375} & \cellcolor{orange!25}29.1  
        & \cellcolor{orange!25}27.841  & 0.832  & \cellcolor{orange!25}0.1844 & \cellcolor{orange!25}1039 & 35.5  \\ 
        % ------------------------------------------
        ~ & 0$\sim$30K
        & \cellcolor{orange!25}24.984  & 0.754  & 0.1967  & \cellcolor{orange!25}1571 & \cellcolor{red!25}\textbf{32.9}
        & \cellcolor{orange!25}31.407  & \cellcolor{orange!25}0.930  & 0.1739 & \cellcolor{red!25}\textbf{375} & \cellcolor{red!25}\textbf{26.4}  
        & 27.839  & 0.832  & 0.1865 & \cellcolor{orange!25}1039 & \cellcolor{red!25}\textbf{30.0}     \\ 
        % ------------------------------------------
    \bottomrule
  \end{tabular}
  }
  \caption{\textbf{Quantitative evaluation on the Mip-NeRF360~\cite{barron2022mipnerf360} reconstructed by Mip-Splatting~\cite{Yu2024MipSplatting}. }
  For the reconstruction of outdoor scenes, Grouping Training with RS yields a larger model size, resulting in the best reconstruction quality. Grouping Training with OPS achieves the fastest reconstruction speed but provides sub-optimal reconstruction quality. For indoor scenes, Grouping Training with OPS continues to offer both the fastest reconstruction speed and the highest reconstruction quality.}
  \label{tab:mip_s_mipnerf360}
  \vspace{-6pt}
\end{table*}

%% file: figures/4_visualization_comp.tex
\begin{figure*}
\centering
\setlength\tabcolsep{1pt}

\resizebox{0.95\textwidth}{!}{ % \footnotesize % \small
\begin{tabular}{cccc}
&
\textbf{\textcolor[rgb]{0.216,0.463,0.616}{Baseline}} & 
\textbf{\textcolor[rgb]{0.812,0.353,0.269}{Baseline with Group Training}} &
\textbf{\textcolor[rgb]{0.435,0.435,0.435}{Ground Truth}} \\

\multirow{1}{*}[9.7ex]{\rotatebox{90}{3DGS}} &
\includegraphics[width=0.32\textwidth]{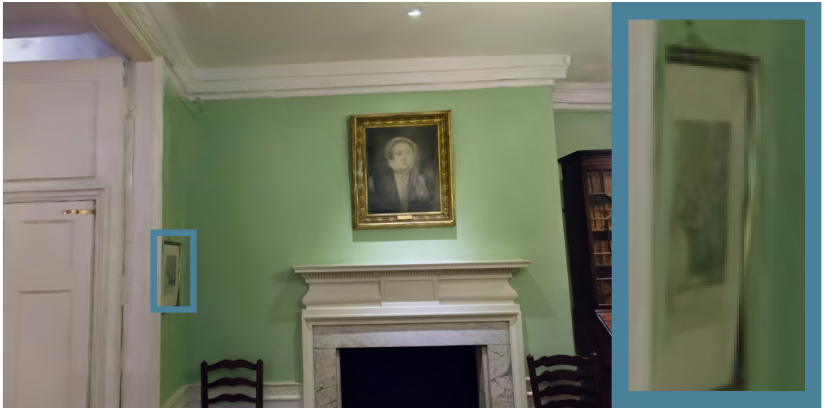} &
\includegraphics[width=0.32\textwidth]{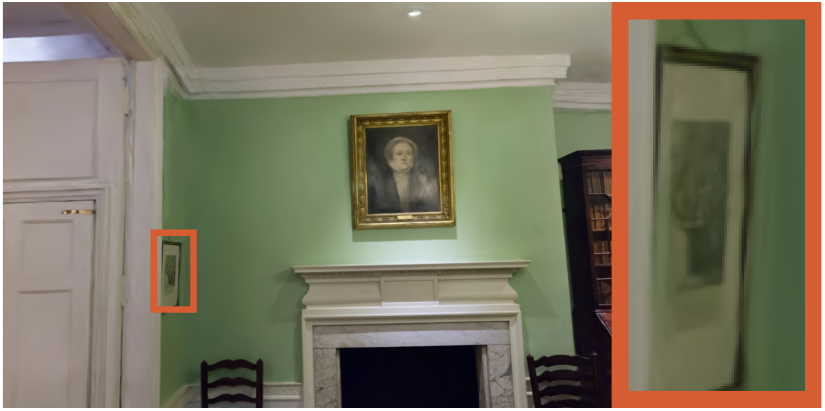} &
\includegraphics[width=0.32\textwidth]{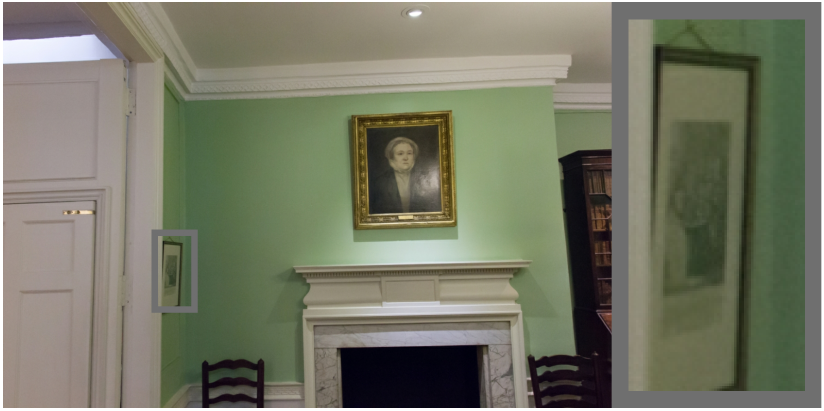} \\
&Time:~26.8 min \quad PSNR:~28.93 dB & Time:~18.6 min \quad PSNR:~29.47 dB & Dr. Johnson~\cite{hedman-2018-deepblending}\\

\multirow{1}{*}[9.7ex]{\rotatebox{90}{3DGS}} &
\includegraphics[width=0.32\textwidth]{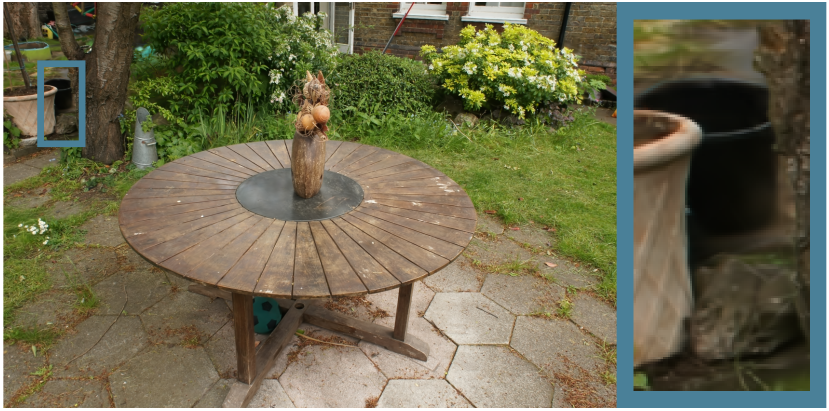} &
\includegraphics[width=0.32\textwidth]{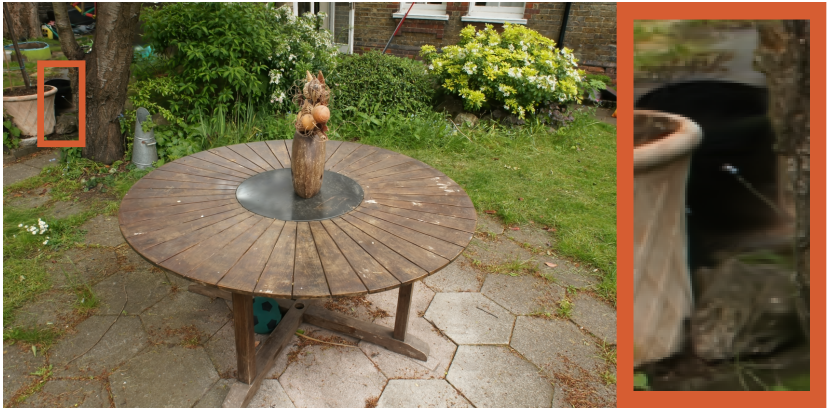} &
\includegraphics[width=0.32\textwidth]{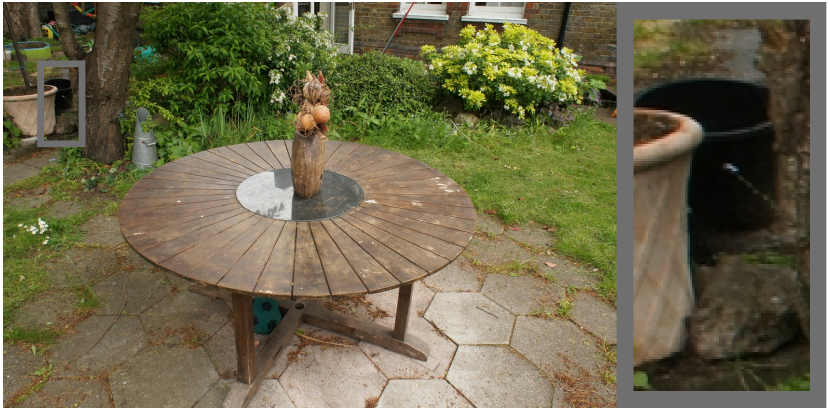} \\
&Time:~35.9 min \quad PSNR:~27.21 dB & Time:~23.2 min \quad PSNR:~27.59 dB & Garden~\cite{barron2022mipnerf360} \\

\multirow{1}{*}[13ex]{\rotatebox{90}{Mip-Splatting}} &
\includegraphics[width=0.32\textwidth]{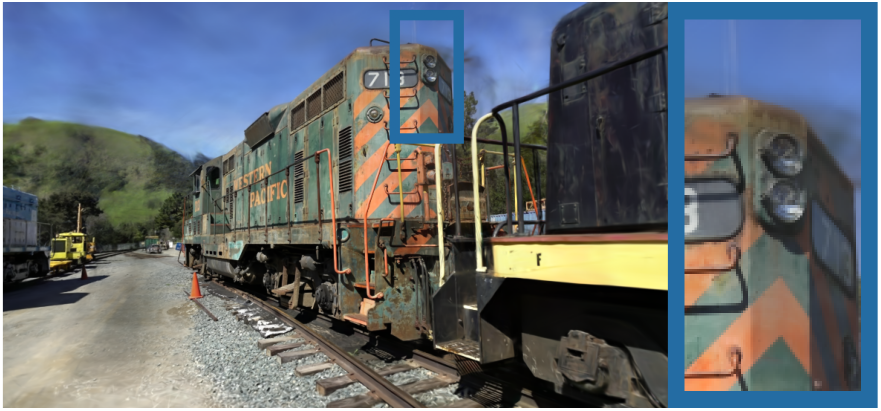} &
\includegraphics[width=0.32\textwidth]{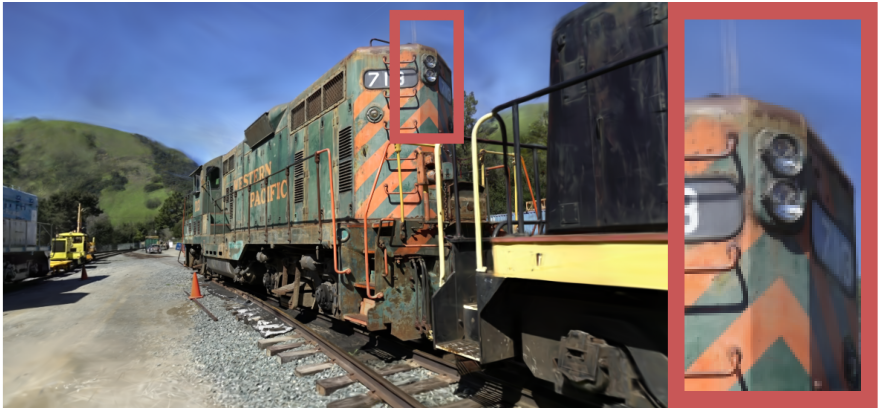} &
\includegraphics[width=0.32\textwidth]{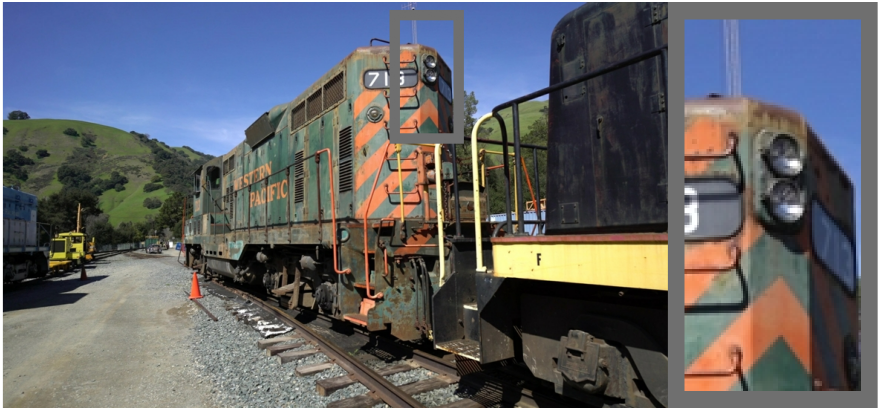} \\
&Time:~19.5 min \quad PSNR:~23.66 dB & Time:~14.7 min \quad PSNR:~24.28 dB & Train~\cite{knapitsch-2017-tanksandtemples} \\

\multirow{1}{*}[13ex]{\rotatebox{90}{Mip-Splatting}} &
\includegraphics[width=0.32\textwidth]{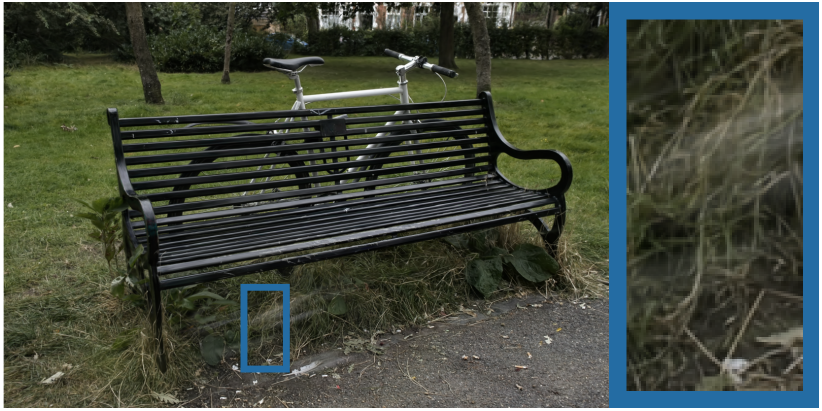} &
\includegraphics[width=0.32\textwidth]{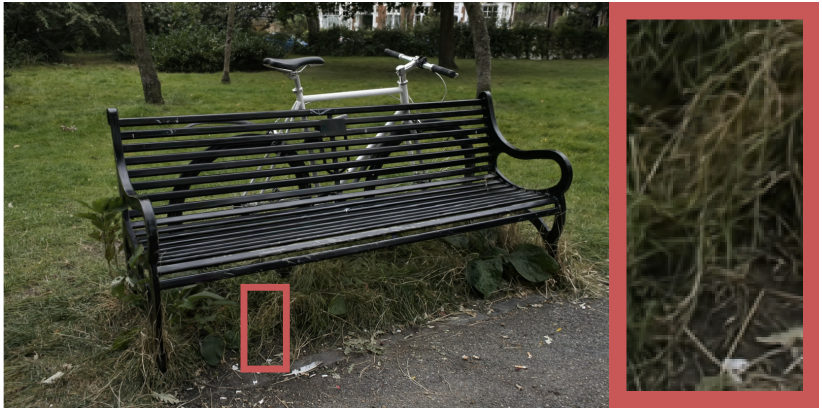} &
\includegraphics[width=0.32\textwidth]{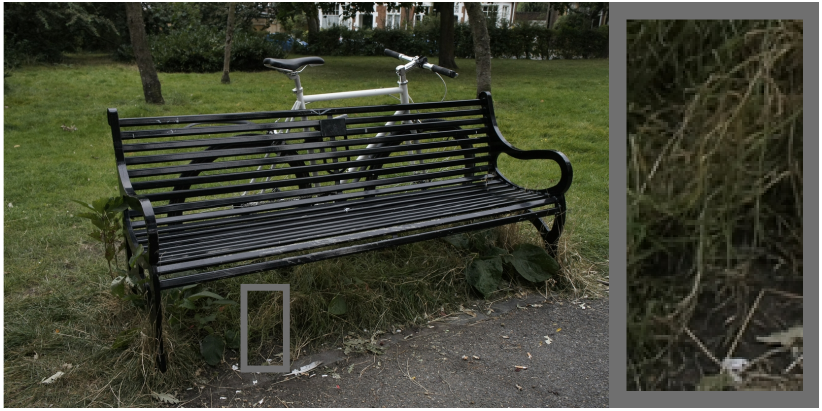} \\
&Time:~54.9 min \quad PSNR:~24.35 dB & Time:~37.7 min \quad PSNR:~25.18 dB & Bicycle~\cite{barron2022mipnerf360} \\

\end{tabular}
}
\caption{
\textbf{Visual Comparison of Grouping Training and Baseline on 3DGS~\cite{kerbl3Dgaussians} and Mip-Splatting~ \cite{Yu2024MipSplatting}. }
    The Grouping Training method accelerates reconstruction by approximately 30\% while achieving more detailed and accurate scene rendering. 
    Notable improvements can be observed in the picture frame in the “Dr. Johnson” scene and the bucket handle in the “Garden” scene, where finer details are captured. 
    Furthermore, Grouping Training significantly reduces floaters in Mip-Splatting, enhancing overall reconstruction fidelity.}
\label{fig:visual-comp}
\vspace{-5pt}
\end{figure*}

%% file: tables/4_exp/ablation_tanks.tex
\begin{table}
  \centering
  \resizebox{\linewidth}{!}{ 
  % \footnotesize 
  \small
  \begin{tabular}{@{}ccccccccc@{}}
    \toprule
      % ~ & \multirow{2}{*}[-0.8ex]{\shortstack{Grouping\\Iterations}} &\multicolumn{4}{c}{Blender} \\
      ~ 
      & \multirow{2}{*}[-0.8ex]{\shortstack{Cyclic\\Resample}} 
      & \multirow{2}{*}[-0.8ex]{\shortstack{Global\\Densify}}
      & \multirow{2}{*}[-0.8ex]{\shortstack{Global\\Optimize}} 
      & \multicolumn{5}{c}{Tanks~\&Temples} \\
      \cmidrule(r){5-9} 
      ~ & ~ & ~ & ~ 
      & PSNR $\uparrow$ & SSIM $\uparrow$ & LPIPS $\downarrow$ & Size $\downarrow$ & Time $\downarrow$ \\
        % ---------------------------------------------------
        \midrule
        3D GS$*$ & -- & -- & -- &
        23.697  & 0.849  & 0.1764  & 434  & 15.0   \\ 
        \midrule
        \multirow{4}{*}{\shortstack{Group\\Training}}
        % \multirow{4}{*}{Ours}
        & \checkmark &            & & 
        23.866  & 0.850  & 0.1749  & 292  & 11.8   \\ 
        & \checkmark & \checkmark & &
        23.769  & 0.849  & 0.1765  & 231  & 11.0   \\ 
        & \checkmark & & \checkmark & 
        23.835  & 0.850  & 0.1754  & 485  & 11.8   \\ 
        & \checkmark & \checkmark & \checkmark & 
        23.853  & 0.850  & 0.1764  & 384  & 11.0   \\
    \bottomrule
     % ---------------------------------------------------
  \end{tabular}
  }
  \caption{\textbf{Ablation Study of Group Training with OPS on Tanks~\&~Temples~\cite{knapitsch-2017-tanksandtemples}. }
  % Cyclic Resampling has the most significant acceleration effect on scene reconstruction, 
  % while Global Densification further enhances reconstruction efficiency by reducing model size. 
  % When Global Densification is enabled, Global Optimization contributes to further improving the model's reconstruction performance.
  Cyclic Resampling provides the most substantial acceleration for scene reconstruction, 
  while Global Densification further improves efficiency by reducing model size. 
  With Global Densification enabled, Global Optimization enhances the reconstruction quality of the model.
  }
  \label{tab:ablation_tanks}
  \vspace{-10pt}
\end{table}

%% file: sec/5_Conclution.tex
\section{Conclusion}
\vspace{-5pt}
This paper presents Group Training, an effective strategy that addresses the computational challenges in 3D Gaussian Splatting by dynamically managing Gaussian primitives in organized groups. Through theoretical analysis and extensive experiments, we demonstrate that our method not only accelerates training by up to 30\% but also improves rendering quality across diverse scenarios. Our opacity-based sampling strategy shows universal compatibility with existing 3DGS frameworks, achieving consistent improvements without compromising model complexity. The success of Group Training reveals the importance of dynamic primitive management in 3DGS optimization, suggesting promising directions for future research, such as adaptive grouping strategies and dynamic sampling mechanisms for more complex scenes. Our work provides valuable insights for future developments in efficient novel view synthesis.

%% file: sec/X_acknowledgment.tex
\section*{Acknowledgements}

This work is supported by the National Key R\&D Program of China (No.2022ZD0119000), 
Hunan Provincial Key R\&D Program of China (No.2024JK2020 and 2024JK2021), 
Hunan Provincial Natural Science Foundation of China (No.2024JJ10027), 
Young Talents of Huxiang (No.Z202433000575), 
and Changsha Science Fund for Distinguished Young Scholars (kq2306002).

%% file: sec/X_suppl.tex
% \clearpage
% \setcounter{page}{1}
% \maketitlesupplementary

% \section{Rationale}
% \label{sec:rationale}
% % 
% Having the supplementary compiled together with the main paper means that:
% % 
% \begin{itemize}
% \item The supplementary can back-reference sections of the main paper, for example, we can refer to \cref{sec:intro};
% \item The main paper can forward reference sub-sections within the supplementary explicitly (e.g. referring to a particular experiment); 
% \item When submitted to arXiv, the supplementary will already included at the end of the paper.
% \end{itemize}
% % 
% To split the supplementary pages from the main paper, you can use \href{https://support.apple.com/en-ca/guide/preview/prvw11793/mac#:~:text=Delete%20a%20page%20from%20a,or%20choose%20Edit%20%3E%20Delete).}{Preview (on macOS)}, \href{https://www.adobe.com/acrobat/how-to/delete-pages-from-pdf.html#:~:text=Choose%20%E2%80%9CTools%E2%80%9D%20%3E%20%E2%80%9COrganize,or%20pages%20from%20the%20file.}{Adobe Acrobat} (on all OSs), as well as \href{https://superuser.com/questions/517986/is-it-possible-to-delete-some-pages-of-a-pdf-document}{command line tools}.

% \clearpage
% \setcounter{page}{1}
\maketitlesupplementary
\appendix

\section{Proof of \textbf{Property 1: Opacity-based Effective Gaussians Densification}}

Under the assumptions of mutual independence between Gaussian attributes and intra-primitive parameter independence, 
the partial derivatives $[\frac{\partial \Delta x}{\partial x_\mathrm{m}}, \frac{\partial \Delta y}{\partial y_\mathrm{m}}]$ for any arbitrary Gaussian admit the following derivation:
\begin{equation}
\resizebox{0.6\columnwidth}{!}{$\displaystyle  % 宽度缩放95%
\begin{aligned}
    \frac{\partial \alpha}{\partial G} &= O + \frac{\partial O}{\partial G}G = O + 0 \cdot G = O,
\end{aligned}$}
\end{equation}
\begin{equation}
\resizebox{0.6\columnwidth}{!}{$\displaystyle  % 宽度缩放百分比
\begin{aligned}
    \frac{\partial L}{\partial x_\mathrm{m}}
        &= 
        \sum\limits_\mathrm{pixel}
        \frac{\partial L}{\partial G^\mathrm{2D}_m}
        \frac{\partial G^\mathrm{2D}_m}{\partial \Delta x}
        \frac{\partial \Delta x}{\partial x_\mathrm{m}}    \\
        &= 
        \sum\limits_\mathrm{pixel}
        o_m\frac{\partial L}{\partial \alpha_m}
        \frac{\partial G^\mathrm{2D}_m}{\partial \Delta x}
        \frac{\partial \Delta x}{\partial x_\mathrm{m}}    \\
        &= 
        o_m
        \sum\limits_\mathrm{pixel}
        \frac{\partial L}{\partial \hat C}
        \frac{\partial \hat C}{\partial \alpha_m}
        \frac{\partial G^\mathrm{2D}_m}{\partial \Delta x}
        \frac{\partial \Delta x}{\partial x_\mathrm{m}},
        \label{eq:dL_dx_s}
\end{aligned}$}
\end{equation}

\begin{equation}
\resizebox{0.6\columnwidth}{!}{$\displaystyle  % 宽度缩放95%
\begin{aligned}
    \frac{\partial L}{\partial y_\mathrm{m}} &= 
        o_m
        \sum\limits_\mathrm{pixel}
        \frac{\partial L}{\partial \hat C}
        \frac{\partial \hat C}{\partial \alpha_m}
        \frac{\partial G^\mathrm{2D}_m}{\partial \Delta y}
        \frac{\partial \Delta y}{\partial y_\mathrm{m}},
        % \label{eq:dL_dy_s}
\end{aligned}$}
\end{equation}
where $[\frac{\partial \Delta x}{\partial x_\mathrm{m}}, \frac{\partial \Delta y}{\partial y_\mathrm{m}}]$ remain constant parameters determined by the resolution $[W,H]$; 
$[\frac{\partial G^\mathrm{2D}_m}{\partial \Delta x},
  \frac{\partial G^\mathrm{2D}_m}{\partial \Delta y}]$ 
derive from the scale, the rotation and the world coordinates of Gaussian primitives (independent of their opacity); and $\frac{\partial L}{\partial \hat{C}}$ represents the loss gradient with respect to the current pixel value.

% {\setlength{\medmuskip}{0.5\medmuskip}
% \begin{equation}
%     \hat{C}     &=
%     \underbrace{\sum\limits_{i=1}^{m-1} \alpha_i c_i \prod\limits_{j=1}^{i-1}{(1-\alpha_j)}}_{\mathrm{Before \, Gaussian}\,m} 
%     + \alpha_m c_m \prod\limits_{j=1}^{m-1}{(1-\alpha_j)} 
%     + \underbrace{\sum\limits_{i=m+1}^{N} \overbrace{\prod\limits_{j=1}^{m-1}{(1-\alpha_j)}}^{\mathrm{Before\,G\,}m} \alpha_i c_i (1-\alpha_m)\overbrace{\prod\limits_{j=m+1}^{i-1}{(1-\alpha_j)}}^{\mathrm{After \, G \, }m}}_{\mathrm{After\,Gaussian\,}m}, 
%     \label{eq:C_hat_G_m}
% \end{equation}
% }
% Given that $\hat{C}$ is formulated as the composite rendering of $N$ Gaussians in~\cref{eq:C_hat_G_m}, then the derivative $\frac{\partial \hat{C}}{\partial \alpha_m}$ admits computation via~\cref{eq:dC_dalpha}.
Given that $\hat{C}$ is formulated as the composite rendering of $N$ Gaussians in~\cref{eq:C_hat_G_m}, the derivative $\frac{\partial \hat{C}}{\partial \alpha_m}$ admits computation via~\cref{eq:dC_dalpha}.
\begin{equation}
\resizebox{0.9\columnwidth}{!}{$\displaystyle  % 宽度缩放95%
\begin{aligned}
    \hat{C}     &=
    \underbrace{\sum\limits_{i=1}^{m-1} \alpha_i c_i \prod\limits_{j=1}^{i-1}{(1-\alpha_j)}}_{\mathrm{Before \, Gaussian}\,m} 
     + 
    \underbrace{\alpha_m c_m \prod\limits_{j=1}^{m-1}{(1-\alpha_j)}}_{\mathrm{Gaussian} \, m} 
    \\& + 
    \underbrace{\sum\limits_{i=m+1}^{N} \overbrace{\prod\limits_{j=1}^{m-1}{(1-\alpha_j)}}^{\mathrm{Before\,G\,}m} \alpha_i c_i (1-\alpha_m)\overbrace{\prod\limits_{j=m+1}^{i-1}{(1-\alpha_j)}}^{\mathrm{After \, G \, }m}}_{\mathrm{After\,Gaussian\,}m}
    \\& + 
    \underbrace{\prod\limits_{i=1}^N(1-\alpha_i)c_\mathrm{bg}}_\mathrm{background}, 
    \label{eq:C_hat_G_m}
\end{aligned}$}
\end{equation}
\begin{equation}
\resizebox{0.99\columnwidth}{!}{$\displaystyle  % 宽度缩放95%
\begin{aligned}
    \frac{\partial \hat{C}}{\partial \alpha_m}
    % &= c_m \prod\limits_{j=1}^{m-1}{(1-\alpha_j)} - \sum\limits_{i=m+1}^{N} \overbrace{\prod\limits_{j=1}^{m-1}{(1-\alpha_j)}}^{\mathrm{Before\,G\,}m} \alpha_i c_i \overbrace{\prod\limits_{j=m+1}^{i-1}{(1-\alpha_j)}}^{\mathrm{After \, G \, }m} 
    % - c_\mathrm{bg} \frac{T_{\mathrm{sat.}}}{1-\alpha_m} 
    % \\&
    = \overbrace{\prod\limits_{j=1}^{m-1}{(1-\alpha_j)}}^{\mathrm{Before\,G\,}m} \left[ c_m  - \sum\limits_{i=m+1}^{N} \alpha_i c_i \overbrace{\prod\limits_{j=m+1}^{i-1}{(1-\alpha_j)}}^{\mathrm{After \, G \, }m} \right]
    - \frac{c_\mathrm{bg} T_{N}}{1-\alpha_m}   % saturation
\end{aligned}
\label{eq:dC_dalpha}
$}
\end{equation}

Subsequently, the mathematical expectation of this derivative is formally established through~\cref{eq:E_dC_dalpha_s}
\begin{equation}
\resizebox{0.99\columnwidth}{!}{$\displaystyle  % 宽度缩放95%
\begin{aligned}
    \mathbb{E} \left[ \frac{\partial \hat{C}}{\partial \alpha_m} \right] 
    &= \overbrace{(1-\alpha_0)^{m-1}}^{\mathrm{Before\,G\,}m} 
    \left\{ 
        c_0 - c_0 \alpha_0   
        \sum\limits_{i=m+1}^{N} 
        \mathbb{E} \left[
            \overbrace{\prod\limits_{j=m+1}^{i-1}{(1-\alpha_j)}}^{\mathrm{After \, G \, }m} 
        \right] 
    \right\} - \frac{c_\mathrm{bg} T_\mathrm{sta.}}{1-\alpha_0}    \\
    % &= \overbrace{(1-\alpha_0)^{m-1}}^{\mathrm{Before\,G\,}m} \left[ c_0 - \alpha_0 c_0 \overbrace{
    % \left( 
    % \prod\limits_{j=m+1}^{i-1}{(1-\alpha_j)}
    % 1
    % +1\cdot(1-\alpha_{m+1})
    % +1\cdot(1-\alpha_{m+1})\cdot(1-\alpha_{m+2})
    % +1\cdot(1-\alpha_{m+1})\cdot(1-\alpha_{m+2})\cdot(1-\alpha_{m+3})
    % +...
    % +1\cdot(1-\alpha_{m+1})...(1-\alpha_{m+3})
    % \right)}^{\mathrm{After \, G \, }m} \right]    \\
    &= (1-\alpha_0)^{m-1}\left[ c_0 - c_0\alpha_0\sum\limits_{i=m+1}^{N}(1-\alpha_0)^{i-m-1} \right]  - \frac{c_\mathrm{bg} T_\mathrm{saturation}}{1-\alpha_0}\\
    % &= (1-\alpha_0)^{m-1}\left[ c_0 - c_0\alpha_0 \frac{1-(1-\alpha_0)^{N-m}}{\alpha_0} \right] \\
    % &= (1-\alpha_0)^{m-1} c_0 \left[ 1 - 1 + (1-\alpha_0)^{N-m} \right] \\
    % &= (1-\alpha_0)^{m-1} c_0 (1-\alpha_0)^{N-m}  \\
    % &= c_0 (1-\alpha_0)^{N-1}  \\
    &= \frac{(c_0 - c_\mathrm{bg}) T_{\mathrm{saturation}}}{ 1-\alpha_0}    \\
    &= \frac{(c_0 - c_\mathrm{bg}) T_{\mathrm{saturation}}}{ 1-\mathbb{E}\left[ o_i\right]\mathbb{E}\left[ G_i\right]},
\end{aligned}
\label{eq:E_dC_dalpha_s}
$}
\end{equation}

\input{tables/X_suppl/sampling_strategy_1}
\input{tables/X_suppl/sampling_strategy_2}

\input{tables/X_suppl/exp_appli}

\input{figures/x_num_G}

\section{Efficiency And Effectiveness For Various Sampling Strategies}

We propose various sampling strategies for Group Training, incorporating Prioritized Sampling based on distinct sampling metrics. 
The sampling probability for each Gaussian primitive $G_i$ is defined as follows:
\begin{equation}
    p_i=\frac{\theta_i}{\sum_{i=1}^N \theta_i }, 
    \label{eq:p_sampling_strategies}
    \vspace{-2pt}
\end{equation}
where $\theta_i$ represents the sampling metrics (opacity~\cite{fan2023lightgaussian}, volume~\cite{liu2025citygaussian} or importance score~\footnote{Based on code: \url{https://github.com/fatPeter/mini-splatting.git}}~\cite{niemeyer2024radsplat,fang2024minisplattingrepresentingscenesconstrained,zhang2024lp}) of Gaussian primitive $G_i$, and $N$ is the total number of Gaussian primitives.
We also evaluated the metric which both Opacity and Volume are considered simultaneously, referred to as the Volume \& Opacity-based method, as applied in \cite{fan2023lightgaussian}. 
The sampling metric $\theta_i$ for Volume \& Opacity-based Prioritized Sampling is computed as follows:
\begin{equation}
    \theta_i=\alpha_i \cdot V_i, 
    \label{eq:theta_v_o}
\end{equation}
where $\alpha_i$ represents the opacity and $V_i$ represents the volume of Gaussian primitive $G_i$.

We conducted experiments using 3D Gaussian Splatting (3DGS) on two datasets: Tanks\&Temples~\cite{knapitsch-2017-tanksandtemples} and Deep Blending~\cite{hedman-2018-deepblending}, both captured with camera-based systems. The comprehensive comparative results are presented in ~\cref{tab:supp_sampling_strategy_1} and ~\cref{tab:supp_sampling_strategy_2}. 
Our results demonstrate that Group Training with Opacity-based Prioritized Sampling (OPS) consistently achieves significant improvements in both reconstruction speed and the quality of 3DGS models. 
Additionally, the reconstructed models exhibit greater compactness, as evidenced by a marked reduction in redundant Gaussian primitives. 

However, volume and importance scores are not the most effective sampling metrics, 
as they fail to differentiate Gaussians that contribute to densification. 
This deficiency leads to abrupt vacancies in the Gaussian space under high sampling rates, 
causing the over-reconstruction and under-reconstruction~\cite{kerbl3Dgaussians}. 
Consequently, this exacerbates Gaussian densification, introducing redundancy between newly densified Gaussians and those already cached. 
% The detailed analysis is provided in ~\cref{sec:mtd_Sampling_Strategies}.
The detailed analysis is provided in Sec.~\textcolor{iccvblue}{3.2}.

\input{figures/x_hitten_FPS}
\input{figures/X_gaussians_uv}
\section{Temporal Evolution of Under-Training Gaussian Primitives}

We visually compare the quantitative differences in under-training Gaussian primitives between Group-Training and 3DGS during scene reconstruction in~\cref{fig:num_G}. 
3DGS with Group-Training reduces the training overhead by avoiding full optimization of all Gaussian primitives. 
Furthermore, during each opacity reset operation, the proposed method retains a higher proportion of geometrically significant primitives compared to baseline. 
These retained elements, despite their low-opacity values, preserve critical structural information that contributes to scene geometry fidelity.

\section{Comparison of Scene Representation Efficiency}

\cref{fig:hitten_FPS} presents a comparative analysis for the "train" scene reconstruction using 3D Gaussian Splatting (3DGS) under baseline conditions versus Group Training. It compares the per-iteration FPS and the number of hitten Gaussians.
And the baseline method required 12.5~minutes to reach a PSNR of 21.985~dB, whereas Group Training with OPS acceleration attained a PSNR of 22.156~dB in just 9.3~minutes. 
These measurements confirm that Group Training consistently accelerates rendering with a substantial reduction in hitten Gaussians count during training. 
Consequently, it demonstrates higher scene representation efficiency by utilizing fewer Gaussian primitives without compromising reconstruction quality.

\section{Distribution of Gaussian Primitives in Imaging Plane Space}
\cref{fig:gaussian_uv} illustrates the projection of rendering Gaussian primitives onto the imaging plane. Our Group Training approach significantly reduces the number of primitives required per image compared to the baseline, without compromising rendering quality, and further improves reconstruction speed.

\section{Methodological Applicability}

We perform comparative validation across two distinct 3DGS architectures: an acceleration-optimized model~\cite{fang2024minisplattingrepresentingscenesconstrained} and a compression-focused LightGaussian~\cite{fan2023lightgaussian}. 

Empirical results demonstrate Group-Training's consistent efficacy across dataset scales, particularly evidenced by reduced temporal overhead in the Blender~\cite{mildenhall2020nerf}, as shown in~\cref{tab:exp_appli}. 
Crucially, our method synergistically integrates with existing acceleration techniques like Mini-Splatting~\cite{fang2024minisplattingrepresentingscenesconstrained}, achieving compounded acceleration gains while providing sustained acceleration for compressed models with concurrent fidelity enhancement.

\section{Detailed Experimental Results for All Scenes}

We present the reconstruction results for all scenes using Group Training with Random Sampling (RS) and Opacity-based Prioritized Sampling (OPS), evaluated on 3D Gaussian Splatting (3DGS)~\cite{kerbl3Dgaussians} and Mip-Splatting~\cite{Yu2024MipSplatting}. The detailed results are provided in ~\cref{tab:supp_3dgs_SyntheticNeRF,tab:supp_3dgs_MipNerf360,tab:supp_3dgs_deep_blending,tab:supp_3dgs_T&T,tab:supp_mip_s_deep_blending,tab:supp_mip_s_MipNerf360,tab:supp_mip_s_SyntheticNeRF,tab:supp_mip_s_T&T}. 

The experimental results demonstrate that Group Training consistently delivers significant improvements in both reconstruction speed and quality across all tests, with the acceleration effect being particularly pronounced on complex datasets. Notably, Group Training with OPS achieves the fastest reconstruction times while maintaining optimal or near-optimal reconstruction quality.

We compare the effects of enabling RS and OPS during the Gaussian densification phase. 
The results indicate that Group Training with RS generates a significantly larger number of Gaussian primitives across all scenarios. 
For example, when reconstructing the “Bicycle” scene using Mip-Splatting, the high density of Gaussian primitives required the use of an NVIDIA A100 GPU for Group Training with RS. 
In contrast, Group Training with OPS produces sparser Gaussian primitives while delivering comparable or even superior reconstruction quality. 
Additionally, the reduced number of Gaussian primitives significantly alleviates the burden on peak memory usage.

\newpage
\input{tables/X_suppl/3dgs_Deep_Blending}
\input{tables/X_suppl/3dgs_mipNeRF360}
\newpage
\input{tables/X_suppl/3dgs_T_T}
\input{tables/X_suppl/3dgs_SyntheticNeRF}

\newpage
\input{tables/X_suppl/mip_s_Deep_Blending}
\input{tables/X_suppl/mip_s_mipNeRF360}

\newpage
\input{tables/X_suppl/mip_s_T_T}
\input{tables/X_suppl/mip_s_SyntheticNeRF}

%% file: tables/X_suppl/sampling_strategy_1.tex
\begin{table}[t]
% \vspace{10pt}
  \centering
  \resizebox{\linewidth}{!}{ % \footnotesize % \small
  \begin{tabular}{@{}ccccccccc@{}}
    \toprule
      ~ & \multirow{2}{*}[-0.8ex]{\shortstack{Sampling\\Strategy}} & \multicolumn{6}{c}{Tanks\&Temples~\cite{knapitsch-2017-tanksandtemples}}  \\
     \cmidrule(r){3-8}
      ~ & ~ & PSNR $\uparrow$ & SSIM $\uparrow$ & LPIPS $\downarrow$ & PM $\downarrow$ & Size $\downarrow$ & Time $\downarrow$ \\
     % ---------------------------------------------------
    \midrule
        3DGS$*$
        ~ & -- & \cellcolor{orange!25}23.730  & \cellcolor{orange!25}0.8491  & \cellcolor{orange!25}0.176  & \cellcolor{orange!25}4.6  & \cellcolor{orange!25}430  & 15.3   \\ 
        \midrule
        \multirow{4}{*}{\shortstack{Group\\Training}}
         & Imp. score 
         & 23.672 & 0.8486 & \cellcolor{red!25}\textbf{0.174} & 5.8 & 593 & 15.7 \\  
         & Vol. & 23.718  & 0.8462  & 0.182  & 5.1  & 493  & 12.4   \\  
         & Opac. & \cellcolor{red!25}\textbf{23.850}  & \cellcolor{red!25}\textbf{0.8500}  & \cellcolor{orange!25}0.176  & \cellcolor{red!25}\textbf{4.5}  & \cellcolor{red!25}\textbf{383}  & \cellcolor{red!25}\textbf{11.0}   \\ 
         & Vol.+Opac. & 23.684  & 0.8475  & 0.179  & 4.8  & 438  & \cellcolor{orange!25}11.9   \\ 
    \bottomrule
     % ---------------------------------------------------
  \end{tabular}
  }
  \caption{\textbf{Quantitative evaluation of training efficiency on the Tanks\&Temples~\cite{knapitsch-2017-tanksandtemples} reconstructed by 3DGS~\cite{kerbl3Dgaussians}. }  
  $*$ indicates that we retrain the model. PM stands for GPU peak memory allocation, with Size in MB and Time in minutes.
  Imp. score = Importance score based, Vol. = Volume-based, Opac. = Opacity-based, Vol.+Opac. = Volume~\&~Opacity-based.}
  \label{tab:supp_sampling_strategy_1}
  % \vspace{10pt}
\end{table}

%% file: tables/X_suppl/sampling_strategy_2.tex
\begin{table}[t]
  \centering
  \resizebox{\linewidth}{!}{ 
  % \footnotesize 
  % \small
  \begin{tabular}{@{}ccccccccc@{}}
    \toprule
      ~ & \multirow{2}{*}[-0.8ex]{\shortstack{Sampling\\Strategy}} & \multicolumn{6}{c}{Deep Blending~\cite{hedman-2018-deepblending}}  \\
     \cmidrule(r){3-8}
      ~ & ~ & PSNR $\uparrow$ & SSIM $\uparrow$ & LPIPS $\downarrow$ & PM $\downarrow$ & Size $\downarrow$ & Time $\downarrow$ \\
     % ---------------------------------------------------
    \midrule
        3DGS$*$
        & -- & 29.503  & 0.9038  & \cellcolor{red!25}\textbf{0.244}  & 7.8  & 677  & 25.2   \\ 
        \midrule
        \multirow{4}{*}{\shortstack{Group\\Training}}
         & Imp. score & 29.589 & \cellcolor{orange!25}0.9051 & 0.246 & 8.5 & 765 & 23.2 \\  
         & Vol. & 29.448  & 0.9036  & 0.251  & 7.5  & 623  & 19.8   \\ 
         & Opac. & \cellcolor{red!25}\textbf{29.768}  & \cellcolor{red!25}\textbf{0.9067}  & \cellcolor{orange!25}0.245  & \cellcolor{red!25}\textbf{6.8}  & \cellcolor{red!25}\textbf{489}  & \cellcolor{red!25}\textbf{17.2}   \\ 
         & Vol.+Opac. & \cellcolor{orange!25}29.619  & 0.9048  & 0.247  & \cellcolor{red!25}7.0  & \cellcolor{orange!25}533  & \cellcolor{orange!25}19.0   \\ 
    \bottomrule
     % ---------------------------------------------------
  \end{tabular}
  }
  \caption{\textbf{Quantitative evaluation of training efficiency on the Deep Blending~\cite{hedman-2018-deepblending} reconstructed by 3DGS~\cite{kerbl3Dgaussians}. }
  Group Training with Opacity-based Prioritised Sampling demonstrates the fastest reconstruction speed and superior performance compared to other sampling strategies.}
    \label{tab:supp_sampling_strategy_2}
    % \vspace{-5pt}
\end{table}

%% file: tables/X_suppl/exp_appli.tex
% \twocolumn[{%
% \renewcommand\twocolumn[1][]{#1}%
% \maketitle
\begin{table*}[!ht]
    \centering
    \resizebox{\textwidth}{!}{ % \small
    % \footnotesize % \small
    \begin{tabular}{@{}ccccccccccccccccccc@{}}
    \toprule
        ~ 
        & \multicolumn{3}{c}{Mip-NeRF360~\cite{barron2022mipnerf360}} 
        & \multicolumn{3}{c}{Tanks\&Temples~\cite{knapitsch-2017-tanksandtemples}} 
        & \multicolumn{3}{c}{Deep Blending~\cite{hedman-2018-deepblending}}  
        & \multicolumn{3}{c}{Blender~\cite{mildenhall2020nerf}} \\
        % & \multicolumn{3}{c}{Mip-NeRF360~[\textcolor{blue}{1}]} 
        % & \multicolumn{3}{c}{Tanks\&Temples~[\textcolor{blue}{13}]} 
        % & \multicolumn{3}{c}{Deep Blending~[\textcolor{blue}{9}]}
        % & \multicolumn{3}{c}{Blender~[\textcolor{blue}{20}]} \\
     \cmidrule(r){2-4} \cmidrule(r){5-7} \cmidrule(r){8-10} \cmidrule(r){11-13}
      ~ & PSNR $\uparrow$ & Time $\downarrow$ & Accel. $\uparrow$
        & PSNR $\uparrow$ & Time $\downarrow$ & Accel. $\uparrow$
        & PSNR $\uparrow$ & Time $\downarrow$ & Accel. $\uparrow$
        & PSNR $\uparrow$ & Time $\downarrow$ & Accel. $\uparrow$ \\
     % ---------------------------------------------------
    \midrule
        3D-GS~
        \cite{kerbl3Dgaussians}
        % [\textcolor{blue}{12}]
          & 27.45  & 26.7  & -- & 23.70  & 15.0  & -- & 29.59  & 23.9  & -- & 33.77  & 6.1  & -- \\ 
        +Group Training
          & 27.56  & 19.6  & 27\% & 23.85  & 11.0  & 27\% & 29.75  & 16.9  & 29\% & 33.81  & \textbf{\textcolor{red}{4.8}}  & 21\% \\ 
        \midrule
        Mini-Splatting~
        \cite{fang2024minisplattingrepresentingscenesconstrained}
        % [\textcolor{blue}{6}]
          & 27.27  & 20.7  & -- & 23.26  & 12.6  &-- & 29.95  & 17.8  & -- & 31.60  & 10.0  & -- \\ 
        +Group Training
          & 27.25  & 17.9  & 13\% & 23.10  & 9.9  & 21\% & 29.85  & 14.7  & 17\% & 31.98  & \textbf{\textcolor{red}{8.4}}  & 16\% \\ 
        \midrule
        LightGaussian~
        \cite{fan2023lightgaussian}
        % [\textcolor{blue}{5}]
          & 27.06  & 27.5  & -- & 23.09  & 16.1  & -- & 27.28  & 25.9  & -- & 32.95  & 6.1  & -- \\ 
        +Group Training
          & 27.34  & 20.5  & 25\% & 23.55  & 11.9  & 26\% & 28.50  & 19.0  & 27\% & 33.18  & \textbf{\textcolor{red}{4.6}}  & 24\% \\ 

    \bottomrule
     % ---------------------------------------------------
  \end{tabular}
  }
    % \vspace{-7pt}
    \captionof{table}{\textbf{Quantitative comparisons on different baselines and datasets.} Group Training with 3DGS achieves faster reconstruction speed than Mini-Splatting across all datasets. Furthermore, Group Training demonstrates \textbf{\textcolor{red}{consistent acceleration effects}} on both 3DGS acceleration model (13\%$\sim$21\% speedup on Mini-Splatting~\cite{fang2024minisplattingrepresentingscenesconstrained}) and compression model (24\%$\sim$27\% speedup on LightGaussian~\cite{fan2023lightgaussian}).Accel. = Acceleration Ratio in training time compared to the baseline. }
    \label{tab:exp_appli}
\end{table*}
% \begin{center}
%     \vspace{-25pt}
    
%     \vspace{-6pt}
% \end{center}%
% % }]

%% file: figures/x_num_G.tex
\begin{figure}[!t]
    \centering
    \includegraphics[trim={0.cm 0 0.cm 0},clip,width=\linewidth]{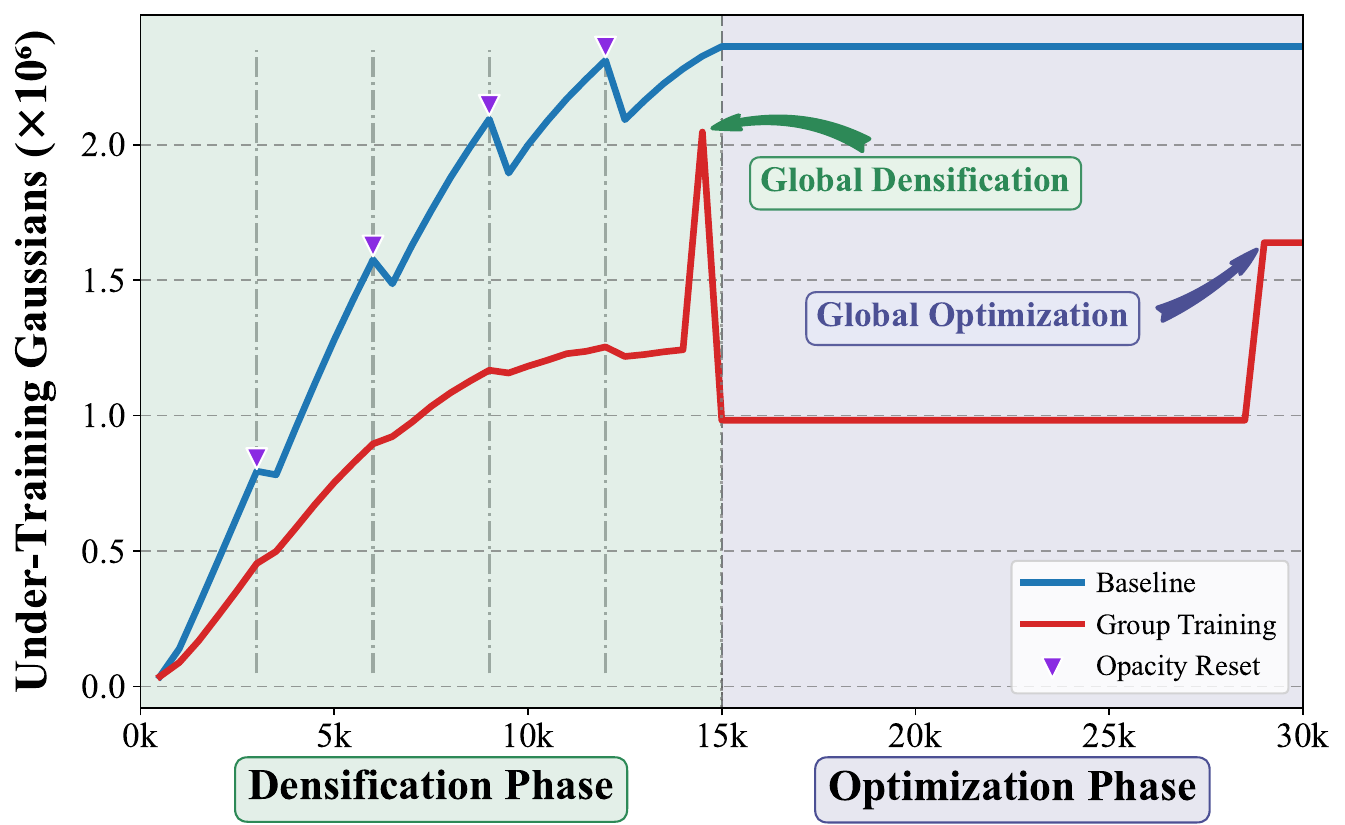}
    \caption{\textbf{Comparsion of Under-Training Gaussian Primitives. }
    Our Group-Training methodology selectively trains a subset of Gaussian primitives, demonstrating enhanced computational efficiency while mitigating loss of potentially critical points during opacity reset operations. 
    }
    \label{fig:num_G}
    % \vspace{-8pt}
\end{figure}

%% file: figures/x_hitten_FPS.tex
\begin{figure}[!t]
    \centering
    \includegraphics[trim={0.cm 0 0.cm 0},clip,width=\linewidth]{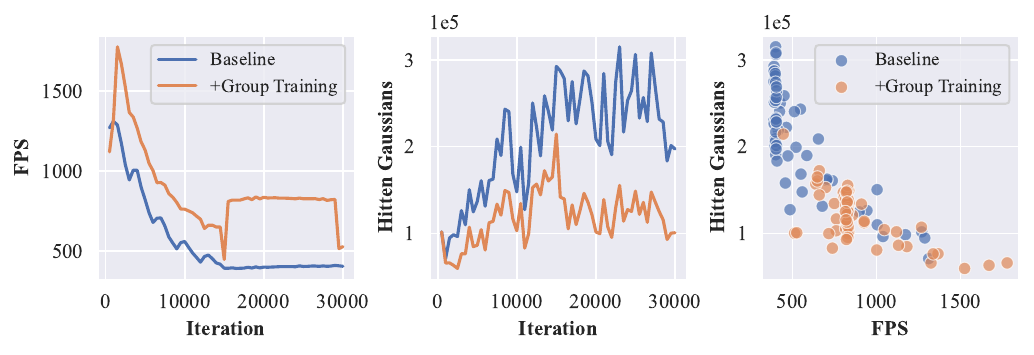}
    \caption{\textbf{Comparsion of Forward Rendering Efficiency. }
    % Our Group-Training methodology selectively trains a subset of Gaussian primitives, demonstrating enhanced computational efficiency while mitigating loss of potentially critical points during opacity reset operations. 
    We measured the number of hit Gaussians and forward rendering FPS throughout the training process. 3DGS with Group Training consistently demonstrated higher FPS and fewer hit Gaussians compared to the baseline method during training.}
    \label{fig:hitten_FPS}
\end{figure}

%% file: figures/X_gaussians_uv.tex
\begin{figure*}[t]
    \begin{subfigure}{0.5\linewidth}
        \centering
        \includegraphics[trim={0.cm 0 0.cm 0}, clip, width=0.95\linewidth]{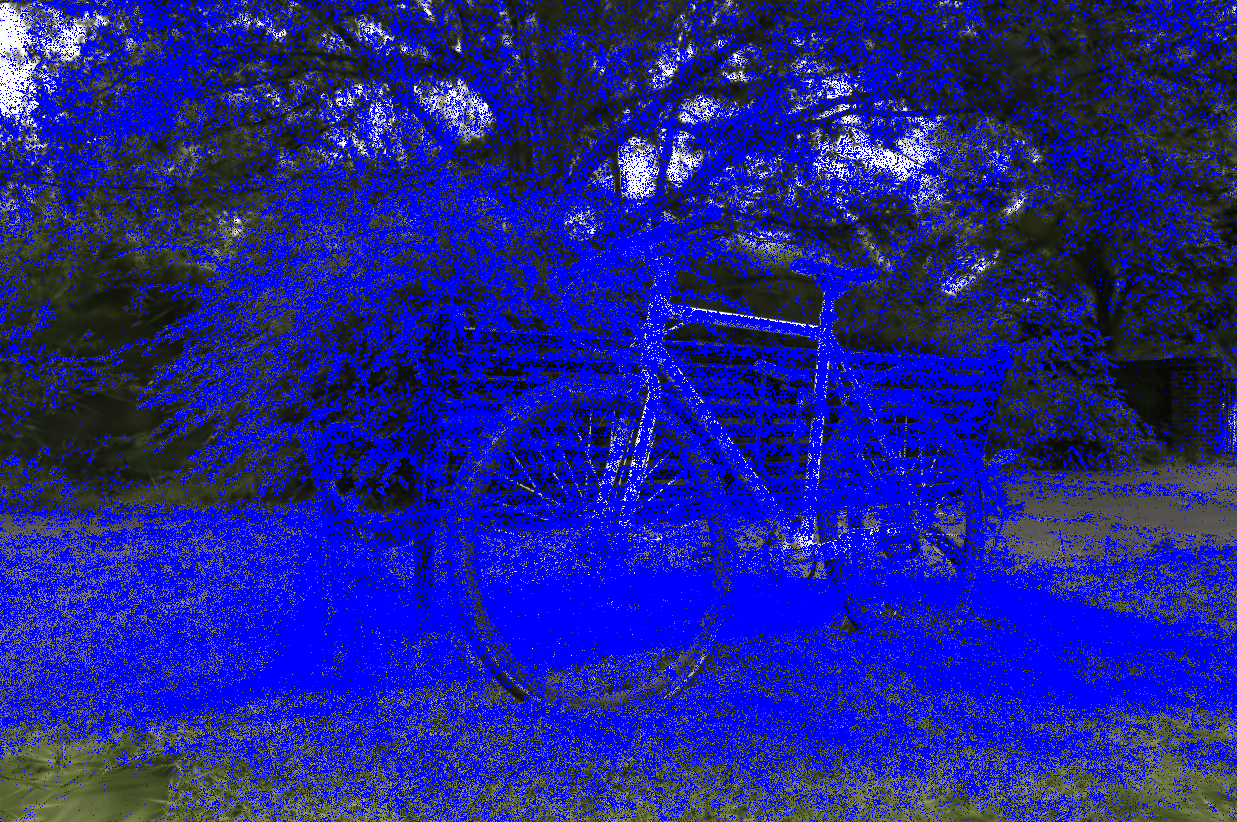}
        \caption{3DGS (PSNR:~25.21dB \quad Time:~34.1min)}
    \end{subfigure}
    \begin{subfigure}{0.5\linewidth}
        \centering
        \includegraphics[trim={0.cm 0 0.cm 0}, clip, width=0.95\linewidth]{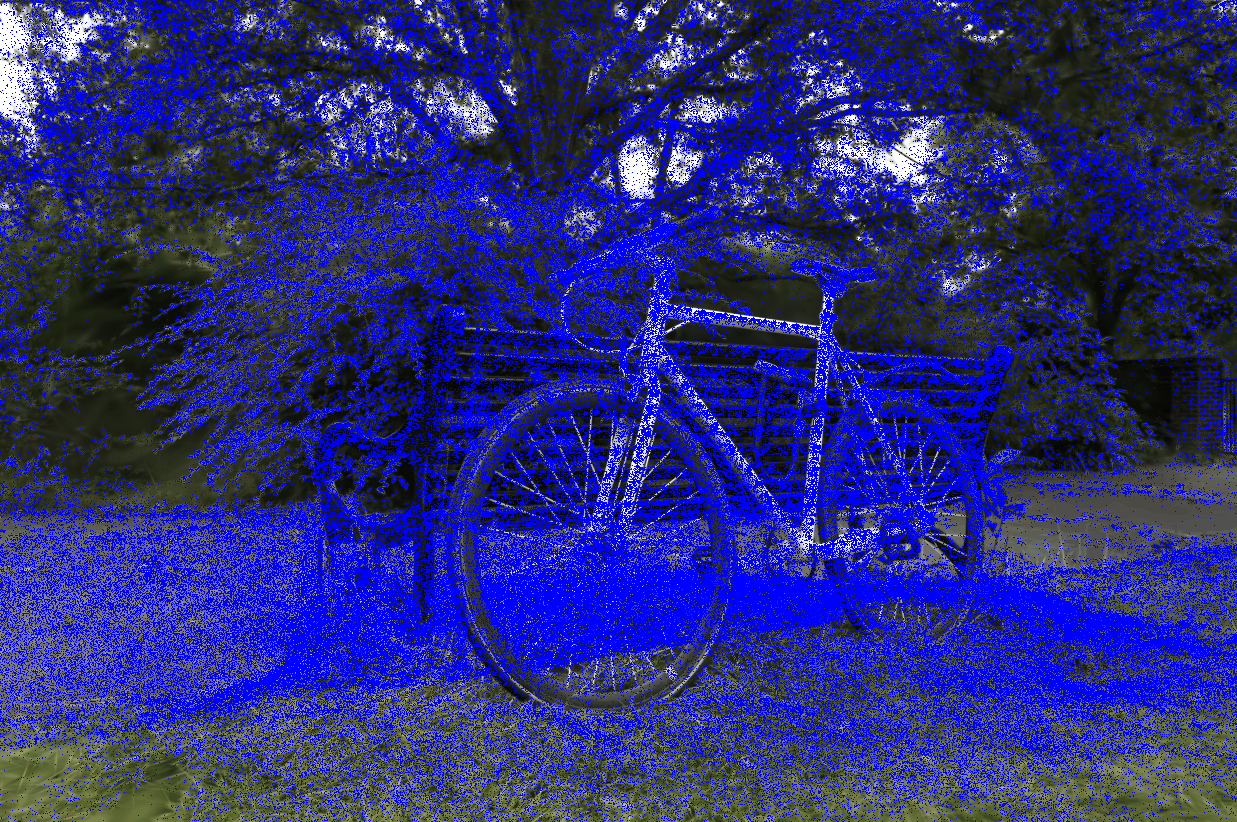}
        \caption{3DGS + Group Training (PSNR:~25.22dB \quad Time:~21.8min)}
    \end{subfigure}
    \caption{
        \textbf{The visual comparison of Gaussian primitive distributions in the imaging plane.}
        We visualize the Gaussian projection information on the imaging plane during images rendering.
        \textbf{Left:} Gaussian distribution on the imaging plane for the "Bicycle" scene~\cite{barron2022mipnerf360}. 
        \textbf{Right:} 3DGS with Group Training achieves comparable rendering quality using fewer Gaussian primitives.
    }
    \label{fig:gaussian_uv}
\end{figure*}

%% file: tables/X_suppl/3dgs_Deep_Blending.tex
\begin{table*}
  \centering
  \resizebox{\textwidth}{!}{ % \footnotesize % \small
  \begin{tabular}{@{}ccccccccccccccc@{}}
    \toprule
      ~ & \multirow{2}{*}[-0.8ex]{\shortstack{Grouping\\Iterations}} & \multicolumn{6}{c}{Dr. Johnson} & \multicolumn{6}{c}{Playroom} \\
     \cmidrule(r){3-8} \cmidrule(r){9-14}
      ~ & ~ & PSNR $\uparrow$ & SSIM $\uparrow$ & LPIPS $\downarrow$ & PM $\downarrow$ & Size $\downarrow$ & Time $\downarrow$   
            & PSNR $\uparrow$ & SSIM $\uparrow$ & LPIPS $\downarrow$ & PM $\downarrow$ & Size $\downarrow$ & Time $\downarrow$ \\
     % ---------------------------------------------------
    \midrule
    
        3D Gaussian Splatting~\cite{kerbl3Dgaussians} 
              & -- & 28.766 & 0.899 & 0.244  & -- & -- & -- & 30.044 & 0.906 & 0.241 & -- & -- & --  \\
        3D Gaussian Splatting$*$
        ~ & -- & 29.190  & 0.901  & 0.2442  & 9.0  & 782  & 26.8  & 29.981  & 0.907  & 0.2431  & 6.4  & 549  & 21.0   \\ 
        \midrule
        \multirow{2}{*}{Group Training w/ RS} 
        ~ & 0$\sim$15K & 28.383  & 0.894  & 0.2517  & 8.8  & 733  & 24.2  & 29.987  & 0.908  & 0.2429  & 6.5  & 554  & 19.5   \\ 
        ~ & 0$\sim$30K & 28.701  & 0.902  & 0.2513  & 8.8  & 734  & 20.3  & 30.133  & 0.912  & 0.2448  & 6.5  & 552  & 16.6   \\ 
        \midrule
        \multirow{2}{*}{Group Training w/ OPS}
        ~ & 0$\sim$15K & 29.287  & 0.903  & 0.2430  & 8.1  & 592  & 21.7  & 30.138  & 0.909  & 0.2439  & 5.6  & 382  & 16.8   \\ 
        ~ & 0$\sim$30K & 29.309  & 0.904  & 0.2451  & 8.1  & 594  & 18.6  & 30.183  & 0.910  & 0.2448  & 5.6  & 380  & 15.1   \\  
    \bottomrule
     % ---------------------------------------------------
  \end{tabular}
  }
  \caption{\textbf{Comprehensive quantitative evaluation results on the DeepBlending~\cite{hedman-2018-deepblending} reconstructed by 3DGS~\cite{kerbl3Dgaussians}. } 
  RS denotes Random Sampling, and OPS denotes Opacity-based Prioritized Sampling. }
  \label{tab:supp_3dgs_deep_blending}
\end{table*}

%% file: tables/X_suppl/3dgs_mipNeRF360.tex
\begin{table*}
  \centering
  \resizebox{\linewidth}{!}{ % \footnotesize % \small
  \begin{tabular}{@{}ccccccccccccccc@{}}
    \toprule
      ~ & ~ & \multirow{2}{*}[-0.8ex]{\shortstack{Grouping\\Iterations}} & \multicolumn{9}{c}{Mip-NeRF360~\cite{barron2022mipnerf360}} \\
     \cmidrule(r){4-12}
      ~ & ~ &~ &  bicycle & flowers & garden & stump & treehill & bonsai & counter & kitchen & room \\
     % ---------------------------------------------------
    \midrule
        \multirow{6}{*}[-1.2ex]{PSNR}
        & 3DGS~\cite{kerbl3Dgaussians} & -- & 
         25.246 & 21.520 & 27.410 & 26.550 & 22.490 & 31.980 & 28.700 & 31.317  & 30.632\\
        & 3DGS$*$ & -- &
        25.205  & 21.484  & 27.397  & 26.620  & 22.514  & 32.202  & 28.980  & 31.222  & 31.377   \\ 
        \cmidrule(r){2-12}
        % \cdashline{2-12}[2pt/1pt]
        & \multirow{2}{*}{\shortstack{Group Training w/ RS}}
        & 0$\sim$15K &
        25.228  & 21.748  & 27.552  & 26.854  & 22.441  & 32.430  & 29.121  & 31.579  & 31.634   \\ 
        & ~ & 0$\sim$30K &
         25.217  & 21.806  & 27.463  & 27.095  & 22.671  & 31.975  & 28.850  & 31.319  & 31.438   \\ 
        \cmidrule(r){2-12}
        % \cdashline{2-12}[2pt/1pt]
        & \multirow{2}{*}{\shortstack{Group Training w/ OPS}}
        & 0$\sim$15K &
        25.260  & 21.751  & 27.434  & 26.809  & 22.402  & 32.312  & 29.031  & 31.539  & 31.699   \\ 
        & ~ & 0$\sim$30K &
         25.219  & 21.741  & 27.418  & 26.830  & 22.522  & 32.205  & 28.973  & 31.425  & 31.744   \\ 
     % ---------------------------------------------------
    \midrule
        \multirow{6}{*}[-1.2ex]{SSIM}
        & 3DGS~\cite{kerbl3Dgaussians} & -- & 
        0.771 & 0.605 & 0.868 & 0.775 & 0.638 & 0.938  & 0.905 & 0.922 & 0.914\\
        & 3DGS$*$ & -- & 
        0.765  & 0.605  & 0.866  & 0.773  & 0.634  & 0.942  & 0.908  & 0.927  & 0.919   \\ 
        \cmidrule(r){2-12}
        % \cdashline{2-12}[2pt/1pt]
        & \multirow{2}{*}{\shortstack{Group Training w/ RS}}
        & 0$\sim$15K &
        0.769  & 0.619  & 0.872  & 0.787  & 0.638  & 0.946  & 0.913  & 0.930  & 0.923   \\ 
        & ~ & 0$\sim$30K &
        0.769  & 0.616  & 0.871  & 0.795  & 0.642  & 0.942  & 0.909  & 0.927  & 0.920   \\ 
        \cmidrule(r){2-12}
        % \cdashline{2-12}[2pt/1pt]
        & \multirow{2}{*}{\shortstack{Group Training w/ OPS}}
        & 0$\sim$15K &
        0.770  & 0.617  & 0.869  & 0.785  & 0.635  & 0.945  & 0.911  & 0.929  & 0.922   \\ 
        & ~ & 0$\sim$30K &
        0.768  & 0.616  & 0.868  & 0.786  & 0.637  & 0.944  & 0.909  & 0.927  & 0.921   \\ 
     % ---------------------------------------------------
    \midrule
        \multirow{6}{*}[-1.2ex]{LPIPS}
        & 3DGS~\cite{kerbl3Dgaussians} & -- & 
        0.205 & 0.336 & 0.103 & 0.210  & 0.317  & 0.205 & 0.204 & 0.129 & 0.220  \\
        & 3DGS$*$ & -- & 
        0.2103  & 0.3355  & 0.1069  & 0.2149  & 0.3240  & 0.2036  & 0.2001  & 0.1261  & 0.2184   \\ 
        \cmidrule(r){2-12}
        % \cdashline{2-12}[2pt/1pt]
        & \multirow{2}{*}{\shortstack{Group Training w/ RS}}
        & 0$\sim$15K &
        0.2094  & 0.3246  & 0.0985  & 0.1989  & 0.3182  & 0.1936  & 0.1909  & 0.1216  & 0.2110   \\ 
        & ~ & 0$\sim$30K &
         0.2225  & 0.3324  & 0.1061  & 0.2032  & 0.3281  & 0.2008  & 0.2008  & 0.1280  & 0.2198   \\ 
        \cmidrule(r){2-12}
        % \cdashline{2-12}[2pt/1pt]
        & \multirow{2}{*}{\shortstack{Group Training w/ OPS}}
        & 0$\sim$15K &
        0.2074  & 0.3262  & 0.1033  & 0.2044  & 0.3203  & 0.1968  & 0.1954  & 0.1245  & 0.2148   \\ 
        & ~ & 0$\sim$30K &
         0.2125  & 0.3307  & 0.1051  & 0.2075  & 0.3226  & 0.1992  & 0.1987  & 0.1272  & 0.2161   \\ 
     % ---------------------------------------------------
    \midrule
        \multirow{5}{*}[-0.6ex]{PM}
        & 3DGS$*$ & -- & 
        11.6  & 7.7  & 11.3  & 9.0  & 7.5  & 7.7  & 6.7  & 8.4  & 8.5   \\ 
        \cmidrule(r){2-12}
        % \cdashline{2-12}[2pt/1pt]
        & \multirow{2}{*}{\shortstack{Group Training w/ RS}}
        & 0$\sim$15K &
        12.1  & 8.8  & 12.0  & 10.9  & 9.0  & 8.4  & 7.1  & 8.5  & 8.8   \\ 
        & ~ & 0$\sim$30K &
         12.0  & 8.8  & 12.0  & 10.9  & 9.0  & 8.4  & 7.2  & 8.5  & 8.7   \\ 
        \cmidrule(r){2-12}
        % \cdashline{2-12}[2pt/1pt]
        & \multirow{2}{*}{\shortstack{Group Training w/ OPS}}
        & 0$\sim$15K &
        11.1  & 8.1  & 10.6  & 9.4  & 8.2  & 7.6  & 6.4  & 7.6  & 8.1   \\ 
        & ~ & 0$\sim$30K &
        11.1  & 8.0  & 10.6  & 9.3  & 8.4  & 7.6  & 6.4  & 7.6  & 8.1   \\
     % ---------------------------------------------------
    \midrule
        \multirow{5}{*}[-0.6ex]{Size}
        & 3DGS$*$ & -- & 
        1450  & 858  & 1391  & 1163  & 896  & 296  & 284  & 428  & 366   \\ 
        \cmidrule(r){2-12}
        % \cdashline{2-12}[2pt/1pt]{2pt}
        & \multirow{2}{*}{\shortstack{Group Training w/ RS}} 
        & 0$\sim$15K &
        1516  & 1019  & 1474  & 1457  & 1106  & 408  & 348  & 439  & 397   \\ 
        & ~ & 0$\sim$30K &
         1491  & 1016  & 1472  & 1452  & 1105  & 407  & 349  & 439  & 390   \\ 
        \cmidrule(r){2-12}
        % \cdashline{2-12}[2pt/1pt]
        ~ & \multirow{2}{*}{\shortstack{Group Training w/ OPS}}
        & 0$\sim$15K &
        1192  & 795  & 1083  & 1078  & 899  & 267  & 225  & 277  & 287   \\ 
        & ~ & 0$\sim$30K &
        1195  & 790  & 1084  & 1060  & 921  & 267  & 225  & 275  & 291   \\ 
     % ---------------------------------------------------
    \midrule
        \multirow{5}{*}[-0.6ex]{Time}
        & 3DGS$*$ & -- & 
        34.1  & 24.0  & 35.9  & 27.2  & 24.0  & 20.5  & 22.8  & 28.3  & 23.7   \\ 
        \cmidrule(r){2-12}
        & \multirow{2}{*}{\shortstack{Group Training w/ RS}} 
        & 0$\sim$15K &
        31.2  & 24.2  & 33.2  & 28.7  & 25.0  & 22.1  & 24.3  & 27.6  & 23.3   \\ 
        & ~ & 0$\sim$30K &
         24.8  & 20.1  & 27.3  & 23.6  & 20.3  & 20.6  & 21.7  & 24.2  & 20.6   \\ 
        \cmidrule(r){2-12}
        & \multirow{2}{*}{\shortstack{Group Training w/ OPS}}
        & 0$\sim$15K &
        26.9  & 21.1  & 27.4  & 23.5  & 22.0  & 18.5  & 20.4  & 22.4  & 20.6   \\ 
        & ~ & 0$\sim$30K &
         21.8  & 17.9  & 23.2  & 19.4  & 18.4  & 17.4  & 18.7  & 20.4  & 18.8   \\ 
    \bottomrule
     % ---------------------------------------------------
  \end{tabular}
  }
  \caption{\textbf{Comprehensive quantitative evaluation results on the Mip-NeRF360~\cite{barron2022mipnerf360} reconstructed by 3DGS~\cite{kerbl3Dgaussians}. } }
  \label{tab:supp_3dgs_MipNerf360}
\end{table*}

%% file: tables/X_suppl/3dgs_T_T.tex
\begin{table*}
  \centering
  \resizebox{\textwidth}{!}{ % \footnotesize % \small
  \begin{tabular}{@{}ccccccccccccccc@{}}
    \toprule
      ~ & \multirow{2}{*}[-0.8ex]{\shortstack{Grouping\\Iterations}} & \multicolumn{6}{c}{Train} & \multicolumn{6}{c}{Truck} \\
     \cmidrule(r){3-8} \cmidrule(r){9-14}
      ~ & ~ & PSNR $\uparrow$ & SSIM $\uparrow$ & LPIPS $\downarrow$ & PM $\downarrow$ & Size $\downarrow$ & Time $\downarrow$   
            & PSNR $\uparrow$ & SSIM $\uparrow$ & LPIPS $\downarrow$ & PM $\downarrow$ & Size $\downarrow$ & Time $\downarrow$ \\
     % ---------------------------------------------------
    \midrule
    
        3D Gaussian Splatting~\cite{kerbl3Dgaussians} 
             & -- & 21.097 & 0.802 & 0.218  & -- & -- & -- & 25.187 & 0.879 & 0.148 & -- & -- & --  \\
        3D Gaussian Splatting$*$
        ~ & -- & 21.985  & 0.815  & 0.2063  & 3.6  & 257  & 12.5  & 25.409  & 0.882  & 0.1464  & 5.5  & 610  & 17.5   \\ 
        \midrule
        \multirow{2}{*}{Group Training w/ RS} 
        ~ & 0$\sim$15K & 22.064  & 0.818  & 0.2031  & 3.9  & 278  & 11.7  & 25.482  & 0.885  & 0.1375  & 6.3  & 714  & 17.5   \\ 
        ~ & 0$\sim$30K & 21.910  & 0.812  & 0.2185  & 3.8  & 273  & 10.1  & 25.495  & 0.884  & 0.1460  & 6.3  & 716  & 14.4   \\ 
        \midrule
        \multirow{2}{*}{Group Training w/ OPS}
        ~ & 0$\sim$15K & 22.159  & 0.818  & 0.2040  & 3.6  & 228  & 10.7  & 25.524  & 0.884  & 0.1411  & 5.5  & 539  & 14.9   \\ 
        ~ & 0$\sim$30K & 22.156  & 0.816  & 0.2104  & 3.6  & 227  & 9.3  & 25.549  & 0.884  & 0.1424  & 5.5  & 540  & 12.6   \\ 
    \bottomrule
     % ---------------------------------------------------
  \end{tabular}
  }
  \caption{\textbf{Comprehensive quantitative evaluation results on the Tanks\&Temples~\cite{knapitsch-2017-tanksandtemples} reconstructed by 3DGS~\cite{kerbl3Dgaussians}. } \\ }
  \label{tab:supp_3dgs_T&T}
\end{table*}

%% file: tables/X_suppl/3dgs_SyntheticNeRF.tex
\begin{table*}
  \centering
  \resizebox{\linewidth}{!}{ % \footnotesize % \small
  \begin{tabular}{@{}cccccccccccccc@{}}
    \toprule
      ~ & ~ & \multirow{2}{*}[-0.8ex]{\shortstack{Grouping\\Iterations}} & \multicolumn{8}{c}{Blender~\cite{mildenhall2020nerf}} \\
     \cmidrule(r){4-11}
      ~ & ~ & ~ &  chair & drumps & ficus & hotdog & lego & materials & mic & ship \\
     % ---------------------------------------------------
    \midrule
        \multirow{6}{*}[-1.2ex]{PSNR}
        & 3DGS~\cite{kerbl3Dgaussians} & -- & 
        33.83 & 26.15 & 34.87 & 37.72 & 35.78 & 30.00 & 35.36 & 30.80 \\
        & 3DGS$*$ & -- &
        35.581  & 26.258  & 35.481  & 38.004  & 36.062  & 30.461  & 36.649  & 31.677  \\ 
        \cmidrule(r){2-11}
        % \cdashline{2-12}[2pt/1pt]
        & \multirow{2}{*}{\shortstack{Group Training w/ RS}}
        & 0$\sim$15K &
        35.736  & 26.273  & 35.494  & 38.242  & 36.580  & 30.675  & 36.842  & 31.829  \\ 
        & ~ & 0$\sim$30K &
        35.637  & 26.224  & 35.436  & 38.142  & 36.441  & 30.588  & 36.786  & 31.765  \\
        \cmidrule(r){2-11}
        % \cdashline{2-12}[2pt/1pt]
        & \multirow{2}{*}{\shortstack{Group Training w/ OPS}}
        & 0$\sim$15K &
        35.688  & 26.270  & 35.487  & 38.145  & 36.435  & 30.569  & 36.719  & 31.800  \\ 
        & ~ & 0$\sim$30K &
         35.623  & 26.227  & 35.467  & 38.017  & 36.335  & 30.452  & 36.654  & 31.692  \\ 
     % ---------------------------------------------------
    \midrule
        \multirow{5}{*}[-1.2ex]{SSIM}
        % & 3DGS~\cite{kerbl3Dgaussians} & -- & 
        % \\
        & 3DGS$*$ & -- & 
        0.988  & 0.955  & 0.987  & 0.985  & 0.983  & 0.960  & 0.993  & 0.906  \\ 
        \cmidrule(r){2-11}
        % \cdashline{2-12}[2pt/1pt]
        & \multirow{2}{*}{\shortstack{Group Training w/ RS}}
        & 0$\sim$15K &
        0.988  & 0.955  & 0.987  & 0.986  & 0.985  & 0.962  & 0.993  & 0.909  \\  
        & ~ & 0$\sim$30K &
        0.988  & 0.956  & 0.987  & 0.987  & 0.985  & 0.963  & 0.993  & 0.910  \\ 
        \cmidrule(r){2-11}
        % \cdashline{2-12}[2pt/1pt]
        & \multirow{2}{*}{\shortstack{Group Training w/ OPS}}
        & 0$\sim$15K &
        0.988  & 0.955  & 0.987  & 0.986  & 0.984  & 0.962  & 0.993  & 0.909  \\ 
        & ~ & 0$\sim$30K &
         0.988  & 0.955  & 0.987  & 0.986  & 0.984  & 0.961  & 0.992  & 0.909  \\ 
     % ---------------------------------------------------
    \midrule
        \multirow{5}{*}[-1.2ex]{LPIPS}
        % & 3DGS~\cite{kerbl3Dgaussians} & -- & 
         % \\
        & 3DGS$*$ & -- & 
        0.0104  & 0.0367  & 0.0118  & 0.0201  & 0.0161  & 0.0370  & 0.0064  & 0.1060  \\ 
        \cmidrule(r){2-11}
        % \cdashline{2-12}[2pt/1pt]
        & \multirow{2}{*}{\shortstack{Group Training w/ RS}}
        & 0$\sim$15K &
        0.0097  & 0.0355  & 0.0117  & 0.0170  & 0.0131  & 0.0340  & 0.0061  & 0.0998  \\ 
        & ~ & 0$\sim$30K &
        0.0107  & 0.0357  & 0.0118  & 0.0184  & 0.0140  & 0.0351  & 0.0063  & 0.1037  \\ 
        \cmidrule(r){2-11}
        % \cdashline{2-12}[2pt/1pt]
        & \multirow{2}{*}{\shortstack{Group Training w/ OPS}}
        & 0$\sim$15K &
        0.0099  & 0.0359  & 0.0118  & 0.0181  & 0.0139  & 0.0356  & 0.0064  & 0.1016  \\ 
        & ~ & 0$\sim$30K &
         0.0102  & 0.0364  & 0.0119  & 0.0189  & 0.0143  & 0.0367  & 0.0065  & 0.1042  \\ 
        % ---------------------------------------------------
    \midrule
        \multirow{5}{*}[-0.6ex]{PM}
        & 3DGS$*$ & -- & 
        3.1  & 2.9  & 2.7  & 2.6  & 2.9  & 2.6  & 2.6  & 2.8  \\ 
        \cmidrule(r){2-11}
        % \cdashline{2-12}[2pt/1pt]
        & \multirow{2}{*}{\shortstack{Group Training w/ RS}}
        & 0$\sim$15K &
        2.9  & 2.8  & 2.6  & 2.7  & 3.0  & 2.6  & 2.6  & 2.8  \\ 
        & ~ & 0$\sim$30K &
        2.9  & 2.8  & 2.6  & 2.7  & 3.0  & 2.6  & 2.6  & 2.8  \\ 
        \cmidrule(r){2-11}
        % \cdashline{2-12}[2pt/1pt]
        & \multirow{2}{*}{\shortstack{Group Training w/ OPS}}
        & 0$\sim$15K &
        2.8  & 2.7  & 2.5  & 2.6  & 2.8  & 2.6  & 2.6  & 2.7  \\ 
        & ~ & 0$\sim$30K &
         2.8  & 2.7  & 2.5  & 2.6  & 2.8  & 2.5  & 2.6  & 2.7  \\ 
    \midrule
        \multirow{5}{*}[-0.6ex]{Size}
        & 3DGS$*$ & -- & 
        116  & 92  & 63  & 44  & 82  & 39  & 46  & 66  \\ 
        \cmidrule(r){2-11}
        % \cdashline{2-12}[2pt/1pt]{2pt}
        & \multirow{2}{*}{\shortstack{Group Training w/ RS}} 
        & 0$\sim$15K &
        88  & 78  & 39  & 46  & 97  & 47  & 42  & 69  \\ 
        & ~ & 0$\sim$30K &
        87  & 78  & 39  & 47  & 97  & 47  & 43  & 69  \\ 
        \cmidrule(r){2-11}
        % \cdashline{2-12}[2pt/1pt]
        ~ & \multirow{2}{*}{\shortstack{Group Training w/ OPS}}
        & 0$\sim$15K &
        59  & 57  & 27  & 31  & 58  & 32  & 30  & 49  \\ 
        & ~ & 0$\sim$30K &
         61  & 58  & 27  & 31  & 58  & 30  & 30  & 49  \\ 
     % ---------------------------------------------------
    \midrule
        \multirow{5}{*}[-0.6ex]{Time}
        & 3DGS$*$ & -- & 
        7.4  & 6.6  & 5.3  & 5.9  & 6.5  & 4.9  & 5.3  & 6.6  \\ 
        \cmidrule(r){2-11}
        & \multirow{2}{*}{\shortstack{Group Training w/ RS}} 
        & 0$\sim$15K &
        6.0  & 5.7  & 4.5  & 6.0  & 6.4  & 5.2  & 4.8  & 6.7  \\ 
        & ~ & 0$\sim$30K &
        5.5  & 5.3  & 4.3  & 5.5  & 5.8  & 4.9  & 4.6  & 6.2  \\ 
        \cmidrule(r){2-11}
        & \multirow{2}{*}{\shortstack{Group Training w/ OPS}}
        & 0$\sim$15K &
        5.3  & 5.3  & 4.1  & 5.2  & 5.3  & 4.6  & 4.5  & 5.8  \\ 
        & ~ & 0$\sim$30K &
        5.1  & 5.0  & 4.0  & 5.0  & 5.0  & 4.5  & 4.2  & 5.5 \\ 
    \bottomrule
     % ---------------------------------------------------
  \end{tabular}
  }
  \caption{\textbf{Comprehensive quantitative evaluation results on the Blender~\cite{mildenhall2020nerf} reconstructed by 3DGS~\cite{kerbl3Dgaussians}. } \\  }
  \label{tab:supp_3dgs_SyntheticNeRF}
\end{table*}

%% file: tables/X_suppl/mip_s_Deep_Blending.tex
% \vspace{10pt}
\begin{table*}
  \centering
  \resizebox{\textwidth}{!}{ % \footnotesize % \small
  \begin{tabular}{@{}cccccccccccccccc@{}}
    \toprule
      ~ & \multirow{2}{*}[-0.8ex]{\shortstack{Grouping\\Iterations}} & \multicolumn{6}{c}{Dr. Johnson} & \multicolumn{6}{c}{Playroom} \\
     \cmidrule(r){3-8} \cmidrule(r){9-14}
      ~ & ~ & PSNR $\uparrow$ & SSIM $\uparrow$ & LPIPS $\downarrow$ & PM $\downarrow$ & Size $\downarrow$ & Time $\downarrow$   
            & PSNR $\uparrow$ & SSIM $\uparrow$ & LPIPS $\downarrow$ & PM $\downarrow$ & Size $\downarrow$ & Time $\downarrow$ \\
     % ---------------------------------------------------
    \midrule
        Mip-Splatting$*$
        ~ & -- & 28.711  & 0.898  & 0.2431  & 10.5  & 981  & 39.3  & 30.005  & 0.907  & 0.2348  & 7.4  & 673  & 30.8   \\ 
        \midrule
        \multirow{2}{*}{Group Training w/ RS} 
        ~ & 0$\sim$15K & 27.957  & 0.892  & 0.2526  & 10.1  & 898  & 34.3  & 29.901  & 0.908  & 0.2335  & 8.0  & 749  & 29.1   \\ 
        ~ & 0$\sim$30K & 28.500  & 0.902  & 0.2486  & 10.1  & 902  & 28.2  & 30.283  & 0.914  & 0.2354  & 8.0  & 748  & 24.7   \\
        \midrule
        \multirow{2}{*}{Group Training w/ OPS}
        ~ & 0$\sim$15K & 29.145  & 0.903  & 0.2393  & 9.4  & 732  & 31.0  & 30.185  & 0.910  & 0.2334  & 6.6  & 520  & 25.2   \\ 
        ~ & 0$\sim$30K & 29.271  & 0.904  & 0.2407  & 9.4  & 734  & 26.0  & 30.305  & 0.911  & 0.2360  & 6.6  & 518  & 22.0   \\ 
    \bottomrule
     % ---------------------------------------------------
  \end{tabular}
  }
  \caption{\textbf{Comprehensive quantitative evaluation results on the DeepBlending~\cite{hedman-2018-deepblending} reconstructed by Mip-Splatting~\cite{Yu2024MipSplatting}. } \\  }
  \label{tab:supp_mip_s_deep_blending}
  % \vspace{1pt}
\end{table*}

%% file: tables/X_suppl/mip_s_mipNeRF360.tex
\vspace{10pt}
\begin{table*}
  \centering
  \resizebox{\linewidth}{!}{ % \footnotesize % \small
  \begin{tabular}{@{}cccccccccccccc@{}}
    \toprule
      ~ & ~ & \multirow{2}{*}[-0.8ex]{\shortstack{Grouping\\Iterations}} & \multicolumn{9}{c}{Mip-NeRF360~\cite{barron2022mipnerf360}} \\
     \cmidrule(r){4-12}
      ~ & ~ &~ &  bicycle & flowers & garden & stump & treehill & bonsai & counter & kitchen & room \\
     % ---------------------------------------------------
    \midrule
        \multirow{5}{*}[-1.1ex]{PSNR}
        & Mip-Splatting$*$ & -- &
        25.535  & 21.753  & 27.603  & 26.874  & 22.304  & 32.301  & 29.214  & 31.803  & 31.740  \\ 
        \cmidrule(r){2-12}
        & \multirow{2}{*}{\shortstack{Group Training w/ RS}}
        & 0$\sim$15K &
        25.664  & 21.960  & 27.726  & 27.124  & 22.371  & 32.299  & 29.291  & 31.719  & 31.756  \\ 
        & ~ & 0$\sim$30K &
        25.784  & 22.257  & 27.881  & 27.446  & 22.711  & 32.381  & 29.084  & 31.510  & 31.682  \\ 
        \cmidrule(r){2-12}
        & \multirow{2}{*}{\shortstack{Group Training w/ OPS}}
        & 0$\sim$15K &
        25.651  & 21.852  & 27.825  & 27.166  & 22.411  & 32.745  & 29.330  & 31.870  & 31.721  \\ 
        & ~ & 0$\sim$30K &
        25.634  & 21.844  & 27.858  & 27.145  & 22.441  & 32.653  & 29.255  & 31.864  & 31.856  \\ 
     % ---------------------------------------------------
    \midrule
        \multirow{5}{*}[-1.1ex]{SSIM}
        & Mip-Splatting$*$ & -- & 
        0.792  & 0.641  & 0.877  & 0.790  & 0.639  & 0.945  & 0.914  & 0.931  & 0.925  \\ 
        \cmidrule(r){2-12}
        & \multirow{2}{*}{\shortstack{Group Training w/ RS}}
        & 0$\sim$15K &
        0.800  & 0.653  & 0.881  & 0.805  & 0.647  & 0.949  & 0.917  & 0.932  & 0.926  \\ 
        & ~ & 0$\sim$30K &
         0.803  & 0.656  & 0.882  & 0.816  & 0.658  & 0.947  & 0.914  & 0.931  & 0.925  \\ 
        \cmidrule(r){2-12}
        & \multirow{2}{*}{\shortstack{Group Training w/ OPS}}
        & 0$\sim$15K &
        0.796  & 0.648  & 0.879  & 0.803  & 0.645  & 0.948  & 0.915  & 0.932  & 0.926  \\ 
        & ~ & 0$\sim$30K &
        0.796  & 0.646  & 0.879  & 0.804  & 0.646  & 0.948  & 0.915  & 0.931  & 0.926  \\ 
     % ---------------------------------------------------
    \midrule
        \multirow{5}{*}[-1.1ex]{LPIPS}
        & Mip-Splatting$*$ & -- & 
        0.1670  & 0.2727  & 0.0950  & 0.1889  & 0.2740  & 0.1881  & 0.1864  & 0.1194  & 0.2011  \\ 
        \cmidrule(r){2-12}
        & \multirow{2}{*}{\shortstack{Group Training w/ RS}}
        & 0$\sim$15K &
        0.1607  & 0.2628  & 0.0895  & 0.1736  & 0.2619  & 0.1823  & 0.1807  & 0.1163  & 0.1995  \\ 
        & ~ & 0$\sim$30K &
         0.1708  & 0.2748  & 0.0933  & 0.1751  & 0.2704  & 0.1874  & 0.1891  & 0.1213  & 0.2050  \\
        \cmidrule(r){2-12}
        & \multirow{2}{*}{\shortstack{Group Training w/ OPS}}
        & 0$\sim$15K &
        0.1655  & 0.2657  & 0.0922  & 0.1782  & 0.2683  & 0.1843  & 0.1847  & 0.1185  & 0.2026  \\
        & ~ & 0$\sim$30K &
        0.1673  & 0.2710  & 0.0929  & 0.1799  & 0.2723  & 0.1861  & 0.1864  & 0.1200  & 0.2029  \\ 
     % ---------------------------------------------------
    \midrule
        \multirow{5}{*}[-1.2ex]{PM}
        & Mip-Splatting$*$ & -- & 
        15.6  & 9.5  & 12.3  & 10.9  & 10.1  & 8.4  & 7.4  & 9.2  & 9.7  \\ 
        \cmidrule(r){2-12}
        & \multirow{2}{*}{\shortstack{Group Training w/ RS}}
        & 0$\sim$15K &
        19.2  & 12.6  & 17.2  & 15.5  & 13.7  & 9.6  & 8.3  & 9.9  & 10.0  \\ 
        & ~ & 0$\sim$30K &
         19.2  & 12.7  & 17.1  & 15.6  & 13.7  & 9.6  & 8.3  & 9.9  & 10.0  \\ 
        \cmidrule(r){2-12}
        & \multirow{2}{*}{\shortstack{Group Training w/ OPS}}
        & 0$\sim$15K &
        15.8  & 10.6  & 13.9  & 12.4  & 11.4  & 8.4  & 7.2  & 8.4  & 9.0  \\ 
        & ~ & 0$\sim$30K &
        15.7  & 10.6  & 13.9  & 12.4  & 11.5  & 8.4  & 7.2  & 8.4  & 9.0  \\ 
     % ---------------------------------------------------
    \midrule
        \multirow{5}{*}[-1.2ex]{Size}
        & Mip-Splatting$*$ & -- & 
        1957  & 1089  & 1475  & 1398  & 1232  & 388  & 364  & 523  & 517  \\ 
        \cmidrule(r){2-12}
        & \multirow{2}{*}{\shortstack{Group Training w/ RS}} 
        & 0$\sim$15K &
        2494  & 1550  & 2194  & 2083  & 1763  & 560  & 491  & 615  & 558  \\ 
        & ~ & 0$\sim$30K &
         2489  & 1570  & 2187  & 2094  & 1764  & 560  & 490  & 610  & 560  \\ 
        \cmidrule(r){2-12}
        & \multirow{2}{*}{\shortstack{Group Training w/ OPS}}
        & 0$\sim$15K &
        1968  & 1230  & 1684  & 1564  & 1410  & 379  & 329  & 388  & 403  \\ 
        & ~ & 0$\sim$30K &
        1961  & 1230  & 1684  & 1562  & 1429  & 378  & 329  & 394  & 412  \\ 
     % ---------------------------------------------------
    \midrule
        \multirow{5}{*}[-1.2ex]{Time}
        & Mip-Splatting$*$ & -- & 
        54.9  & 35.4  & 49.5  & 39.3  & 37.3  & 28.6  & 31.9  & 39.0  & 33.6  \\ 
        \cmidrule(r){2-12}
        & \multirow{2}{*}{\shortstack{Group Training w/ RS}} 
        & 0$\sim$15K &
        43.7  & 39.9  & 57.7  & 46.3  & 42.1  & 32.2  & 34.1  & 40.3  & 32.8  \\ 
        & ~ & 0$\sim$30K &
         35.1  & 32.5  & 45.3  & 37.0  & 34.2  & 28.8  & 30.0  & 34.2  & 28.9  \\ 
        \cmidrule(r){2-12}
        & \multirow{2}{*}{\shortstack{Group Training w/ OPS}}
        & 0$\sim$15K &
        47.9  & 34.3  & 47.7  & 37.4  & 36.2  & 27.2  & 28.7  & 32.0  & 28.5  \\ 
        & ~ & 0$\sim$30K &
        37.7  & 28.3  & 38.8  & 30.0  & 29.5  & 24.9  & 26.1  & 28.5  & 25.9 \\ 
    \bottomrule
     % ---------------------------------------------------
  \end{tabular}
  }
  \caption{\textbf{Comprehensive quantitative evaluation results on the Mip-NeRF360~\cite{barron2022mipnerf360} reconstructed by Mip-Splatting~\cite{Yu2024MipSplatting}.} \\  }
  \label{tab:supp_mip_s_MipNerf360}
  \vspace{10pt}
\end{table*}

%% file: tables/X_suppl/mip_s_T_T.tex
\begin{table*}
  \centering
  \resizebox{\textwidth}{!}{ % \footnotesize % \small
  \begin{tabular}{@{}ccccccccccccccc@{}}
    \toprule
      ~ & \multirow{2}{*}[-0.8ex]{\shortstack{Grouping\\Iterations}} & \multicolumn{6}{c}{Train} & \multicolumn{6}{c}{Truck} \\
     \cmidrule(r){3-8} \cmidrule(r){9-14}
      ~ & ~ & PSNR $\uparrow$ & SSIM $\uparrow$ & LPIPS $\downarrow$ & PM $\downarrow$ & Size $\downarrow$ & Time $\downarrow$   
            & PSNR $\uparrow$ & SSIM $\uparrow$ & LPIPS $\downarrow$ & PM $\downarrow$ & Size $\downarrow$ & Time $\downarrow$ \\
     % ---------------------------------------------------
    \midrule
        Mip-Splatting$*$
        ~ & -- & 21.783  & 0.826  & 0.1892  & 4.4  & 351  & 19.5  & 25.714  & 0.893  & 0.1232  & 6.8  & 767  & 26.5   \\ 
        \midrule
        \multirow{2}{*}{Group Training w/ RS} 
        ~ & 0$\sim$15K & 22.004  & 0.829  & 0.1861  & 4.8  & 405  & 18.9  & 25.901  & 0.896  & 0.1146  & 9.2  & 1123  & 30.3   \\ 
        ~ & 0$\sim$30K & 22.358  & 0.829  & 0.1975  & 4.7  & 403  & 16.3  & 25.934  & 0.896  & 0.1221  & 9.2  & 1119  & 24.4   \\ 
        \midrule
        \multirow{2}{*}{Group Training w/ OPS}
        ~ & 0$\sim$15K & 21.994  & 0.830  & 0.1859  & 4.4  & 346  & 17.4  & 25.921  & 0.896  & 0.1178  & 7.5  & 874  & 25.9   \\ 
        ~ & 0$\sim$30K & 22.167  & 0.829  & 0.1915  & 4.4  & 348  & 14.7  & 25.991  & 0.895  & 0.1205  & 7.6  & 876  & 21.1   \\ 
    \bottomrule
     % ---------------------------------------------------
  \end{tabular}
  }
  \caption{\textbf{Comprehensive quantitative evaluation results on the Tanks\&Temples~\cite{knapitsch-2017-tanksandtemples} reconstructed by Mip-Splatting~\cite{Yu2024MipSplatting}. } \\  }
  \label{tab:supp_mip_s_T&T}
\end{table*}

%% file: tables/X_suppl/mip_s_SyntheticNeRF.tex
\begin{table*}
  \centering
  \resizebox{\linewidth}{!}{ % \footnotesize % \small
  \begin{tabular}{@{}cccccccccccccc@{}}
    \toprule
      ~ & ~ 
      & \multirow{2}{*}[-0.8ex]{\shortstack{Grouping\\Iterations}} 
      & \multicolumn{8}{c}{Blender~\cite{mildenhall2020nerf}} \\
        \cmidrule(r){4-11}
      ~ & ~ & ~ &  chair & drumps & ficus & hotdog & lego & materials & mic & ship \\
     % ---------------------------------------------------
    \midrule
        \multirow{5}{*}[-1.1ex]{PSNR}
        & Mip-Splatting$*$ & -- &
        35.773  & 26.357  & 35.890  & 38.267  & 36.354  & 30.645  & 36.934  & 31.738  \\ 
        \cmidrule(r){2-11}
        & \multirow{2}{*}{\shortstack{Group Training w/ RS}}
        & 0$\sim$15K &
        36.078  & 26.367  & 35.930  & 38.474  & 36.883  & 30.839  & 37.152  & 31.902  \\ 
        & ~ & 0$\sim$30K &
        35.910  & 26.293  & 35.865  & 38.376  & 36.860  & 30.780  & 36.968  & 31.931  \\ 
        \cmidrule(r){2-11}
        & \multirow{2}{*}{\shortstack{Group Training w/ OPS}}
        & 0$\sim$15K &
        35.981  & 26.369  & 35.912  & 38.396  & 36.748  & 30.756  & 37.067  & 31.823  \\ 
        & ~ & 0$\sim$30K &
         35.958  & 26.363  & 35.914  & 38.326  & 36.739  & 30.704  & 37.049  & 31.824  \\ 
     % ---------------------------------------------------
    \midrule
        \multirow{5}{*}[-1.1ex]{SSIM}
        & Mip-Splatting$*$ & -- & 
        0.988  & 0.956  & 0.988  & 0.986  & 0.984  & 0.961  & 0.993  & 0.907  \\ 
        \cmidrule(r){2-11}
        & \multirow{2}{*}{\shortstack{Group Training w/ RS}}
        & 0$\sim$15K &
        0.989  & 0.956  & 0.988  & 0.987  & 0.986  & 0.963  & 0.993  & 0.909  \\ 
        & ~ & 0$\sim$30K &
        0.989  & 0.957  & 0.988  & 0.987  & 0.986  & 0.964  & 0.993  & 0.911  \\ 
        \cmidrule(r){2-11}
        & \multirow{2}{*}{\shortstack{Group Training w/ OPS}}
        & 0$\sim$15K &
        0.989  & 0.956  & 0.988  & 0.987  & 0.985  & 0.963  & 0.993  & 0.909  \\ 
        & ~ & 0$\sim$30K &
         0.989  & 0.956  & 0.988  & 0.987  & 0.985  & 0.963  & 0.993  & 0.909  \\ 
     % ---------------------------------------------------
    \midrule
        \multirow{5}{*}[-1.1ex]{LPIPS}
        & Mip-Splatting$*$ & -- & 
        0.0109  & 0.0366  & 0.0111  & 0.0186  & 0.0150  & 0.0359  & 0.0062  & 0.1022  \\ 
        \cmidrule(r){2-11}
        & \multirow{2}{*}{\shortstack{Group Training w/ RS}}
        & 0$\sim$15K &
        0.0101  & 0.0355  & 0.0110  & 0.0163  & 0.0126  & 0.0330  & 0.0059  & 0.0980  \\ 
        & ~ & 0$\sim$30K &
        0.0110  & 0.0358  & 0.0111  & 0.0173  & 0.0132  & 0.0341  & 0.0060  & 0.0998  \\ 
        \cmidrule(r){2-11}
        & \multirow{2}{*}{\shortstack{Group Training w/ OPS}}
        & 0$\sim$15K &
        0.0105  & 0.0360  & 0.0111  & 0.0169  & 0.0133  & 0.0343  & 0.0060  & 0.0997  \\ 
        & ~ & 0$\sim$30K &
         0.0105  & 0.0359  & 0.0111  & 0.0171  & 0.0134  & 0.0349  & 0.0061  & 0.1001  \\ 
     % ---------------------------------------------------
    \midrule
        \multirow{5}{*}[-1.0ex]{PM}
        & Mip-Splatting$*$ & -- & 
        3.0  & 3.0  & 2.7  & 2.7  & 2.9  & 2.6  & 2.8  & 2.9  \\ 
        \cmidrule(r){2-11}
        & \multirow{2}{*}{\shortstack{Group Training w/ RS}}
        & 0$\sim$15K &
        3.3  & 3.3  & 2.9  & 2.9  & 3.3  & 2.8  & 3.0  & 3.4  \\ 
        & ~ & 0$\sim$30K &
        3.3  & 3.4  & 2.9  & 2.9  & 3.3  & 2.8  & 3.0  & 3.4  \\ 
        \cmidrule(r){2-11}
        & \multirow{2}{*}{\shortstack{Group Training w/ OPS}}
        & 0$\sim$15K &
        3.1  & 3.1  & 2.7  & 2.7  & 3.1  & 2.7  & 2.9  & 3.1  \\ 
        & ~ & 0$\sim$30K &
         3.1  & 3.1  & 2.7  & 2.7  & 3.1  & 2.7  & 2.9  & 3.1  \\ 
     % ---------------------------------------------------
    \midrule
        \multirow{5}{*}[-1.0ex]{Size}
        & Mip-Splatting$*$ & -- & 
        90  & 98  & 51  & 51  & 76  & 40  & 64  & 83  \\ 
        \cmidrule(r){2-11}
        & \multirow{2}{*}{\shortstack{Group Training w/ RS}} 
        & 0$\sim$15K &
        141  & 146  & 79  & 70  & 136  & 71  & 94  & 146  \\ 
        & ~ & 0$\sim$30K &
        141  & 147  & 80  & 71  & 136  & 71  & 94  & 146  \\ 
        \cmidrule(r){2-11}
        & \multirow{2}{*}{\shortstack{Group Training w/ OPS}}
        & 0$\sim$15K &
        109  & 115  & 57  & 55  & 98  & 51  & 72  & 109  \\ 
        & ~ & 0$\sim$30K &
         109  & 115  & 57  & 55  & 99  & 51  & 72  & 109  \\ 
    \midrule
        \multirow{5}{*}[-1.0ex]{Time}
        & Mip-Splatting$*$ & -- & 
        9.2  & 8.8  & 6.2  & 8.4  & 8.8  & 6.2  & 9.3  & 10.7  \\ 
        \cmidrule(r){2-11}
        & \multirow{2}{*}{\shortstack{Group Training w/ RS}} 
        & 0$\sim$15K &
        9.6  & 9.7  & 6.9  & 8.8  & 9.8  & 7.3  & 9.8  & 12.1  \\ 
        & ~ & 0$\sim$30K &
        8.5  & 8.7  & 6.4  & 8.1  & 8.8  & 6.9  & 8.5  & 10.8  \\ 
        \cmidrule(r){2-11}
        & \multirow{2}{*}{\shortstack{Group Training w/ OPS}}
        & 0$\sim$15K &
        8.6  & 8.8  & 6.2  & 7.8  & 8.6  & 6.6  & 8.4  & 10.4  \\ 
        & ~ & 0$\sim$30K &
         7.8  & 8.0  & 6.0  & 7.3  & 7.7  & 6.3  & 7.6  & 9.6 \\ 
    \bottomrule
     % ---------------------------------------------------
  \end{tabular}
  }
  \caption{\textbf{Comprehensive quantitative evaluation results on the Blender~\cite{mildenhall2020nerf} reconstructed by Mip-Splatting~\cite{Yu2024MipSplatting}. } \\  }
  \label{tab:supp_mip_s_SyntheticNeRF}
\end{table*}